\pdfoutput=1

\documentclass[11pt]{article}

\usepackage{acl}

\usepackage{times}
\usepackage{latexsym}
\usepackage{booktabs}
\usepackage{makecell}
\usepackage{multirow}
\usepackage{subfigure}

\usepackage[LGR,T1]{fontenc}

\usepackage[utf8]{inputenc}

\usepackage{microtype}

\usepackage{inconsolata}
\usepackage{booktabs} 
\usepackage{tablefootnote}

\newcommand{\method}{\textsc{mEdIT}\space}
\newcommand{\methodnosp}{\textsc{mEdIT}}
\newcommand{\coedit}{\textsc{CoEdIT}\space}
\newcommand{\coeditnosp}{\textsc{CoEdIT}}

\newcommand{\ZH}[1]{
\begin{CJK*}{UTF8}{gbsn}
#1
\end{CJK*}
}

\newcommand{\JA}[1]{
\begin{CJK*}{UTF8}{goth}
#1
\end{CJK*}
}

\newcommand{\KO}[1]{
\begin{CJK*}{UTF8}{mj}
#1
\end{CJK*}
}

\usepackage{microtype}

\usepackage{tabularx}
\usepackage{graphicx}
\usepackage{xspace}
\usepackage{CJKutf8}

\usepackage{arabtex}
\usepackage{utf8}
\setcode{utf8}
\vocalize

\usepackage[framemethod=TikZ]{mdframed}

\title{\methodnosp: Multilingual Text Editing via Instruction Tuning}

\author{Vipul Raheja$^{1}$\ \hspace{0.5cm} Dimitris Alikaniotis$^1$\ \hspace{0.5cm} Vivek Kulkarni$^1$\ \\  \textbf{Bashar Alhafni}$^{2*}$\ \hspace{0.5cm} \textbf{Dhruv Kumar}$^1$ \\
$^1$Grammarly \hspace{0.5cm} $^2$New York University Abu Dhabi \\
\texttt{firstname.lastname@grammarly.com}}

\begin{document}
\setarab
\maketitle

\begingroup\def\thefootnote{*}\footnotetext{Work done during an internship at Grammarly.}\endgroup

\begin{abstract}
We introduce \methodnosp, a multilingual extension to \coedit -- the recent state-of-the-art text editing models for writing assistance. 
\method models are trained by fine-tuning multilingual large, pre-trained language models (LLMs) via instruction tuning. They are designed to take instructions from the user specifying the attributes of the desired text in the form of natural language instructions, such as \emph{Grammatik korrigieren} (German) or \hspace{-0.8cm} \KO{이 \hspace{0.005cm} 텍스트를 \hspace{0.005cm} 단순화} (Korean).
We build \method by curating data from multiple publicly available human-annotated text editing datasets for three text editing tasks (Grammatical Error Correction (GEC), Text Simplification, and Paraphrasing) across diverse languages belonging to six different language families.
We detail the design and training of \method models and demonstrate their strong performance on many multilingual text editing benchmarks against other multilingual LLMs.
We also find that \method generalizes effectively to new languages over multilingual baselines. We publicly release our data, code, and trained models.\footnote{\url{https://github.com/vipulraheja/medit}}
\end{list} 

\end{abstract}

\section{Introduction}
Large language models (LLMs) have made remarkable progress toward generating fluent and coherent text in a wide variety of tasks and domains to support writing assistance (\citealt{brown2020language, openai2023gpt4, touvron2023llama}; \textit{inter alia}). In particular, LLMs have been adapted to perform many complex text editing tasks like GEC \cite{wu2023chatgpt, coyne2023analysis, fang2023chatgpt}, text simplification~\cite{baez-saggion-2023-lsllama,saggion-etal-2022-findings}, paraphrasing \cite{witteveen-andrews-2019-paraphrasing, niu-etal-2021-unsupervised}, and formality and tone rewriting \cite{reif-etal-2022-recipe, luo-etal-2023-prompt}, among others. However, most of these works are restricted to single tasks, with few works adapting LLMs to perform high-quality text editing across multiple tasks \cite{schick2023peer, raheja2023coedit, laban2023beyond}. A lot of these improvements have been driven by fine-tuning large language models (LLMs) with task-specific instruction tuning, resulting in remarkable zero-shot generalization abilities (\citealt{sanh2022multitask, ouyang2022training, chung2022scaling}; \textit{inter alia}).

\begin{figure}
\mdfdefinestyle{mdframedstyle}{innertopmargin=10pt,innerbottommargin=10pt,roundcorner=5pt}
\begin{mdframed}[style=mdframedstyle]
\small

\textbf{Multilingual Editing}   \hfill \vspace{0.1cm} \break
\texttt{\textcolor[HTML]{DC3220}{\textbf{Parafrasee la oración:} Hoy iré a la escuela a estudiar español.}}    \hfill \break
\texttt{\textcolor[HTML]{005AB5}{Hoy asistiré a la escuela para aprender español.}}\hfill \break 

\texttt{\textcolor[HTML]{DC3220}{\hspace{-0.8cm}\textbf{\KO{문법 \hspace{0.005cm} 오류 \hspace{0.005cm} 수정:}}}} \texttt{\textcolor[HTML]{DC3220}{\hspace{-1.005cm}\KO{우리는 \hspace{0.005cm} 어제 \hspace{0.005cm} 행사에서 \hspace{0.005cm} 즐거운 \hspace{0.005cm} 시간을 보내겠습니다.}}} \hfill \break
\texttt{\textcolor[HTML]{005AB5}{\hspace{-0.8cm}\KO{우리는 \hspace{0.005cm} 어제 \hspace{0.005cm} 행사에서 \hspace{0.005cm} 즐거운 \hspace{0.005cm} 시간을 보냈습니다.}}} \hfill \break
\textbf{Cross-lingual Editing} \hfill \vspace{0.15cm} \break
\vspace{0.1cm}
\textcolor[HTML]{DC3220}{\hspace{-0.4cm}\textbf{\JA{文を簡略化してください:}}} 
\texttt{\textcolor[HTML]{DC3220}{Meteorologists often regard a storm surge as the most treacherous facet of a hurricane.}} \hfill \break
\texttt{\textcolor[HTML]{005AB5}{A storm surge is considered a hurricane's most dangerous aspect.}}\hfill 


\end{mdframed}
\caption{\textbf{Examples illustrating multilingual and cross-lingual text editing.} The editing instructions are described in bold. Note that the input and output texts are always in the same language. The monolingual vs. cross-lingual setting is determined by comparing the language of the edit instruction to the language of the input text.}
\label{fig:main_examples}
\end{figure}

At the same time, significant research effort has been dedicated to leveraging and enhancing the multilingual capabilities of LLMs \cite{lin-etal-2022-shot}. These abilities can be improved using methods such as continued pre-training with abundant monolingual data \cite{yang-etal-2023-BigTranslate, chinese-llama-alpaca} or language-specific instruction-tuning \cite{zhu2023extrapolating,li2023bactrianx}. 
However, in the case of continued pre-training, the lack of high-quality web-scale data often restricts the ability to improve LLMs capabilities in less-represented languages in the same way that English data can be expanded. Moreover, while numerous multilingual instruction-tuned models have been developed \cite{muennighoff-etal-2023-crosslingual, workshop2023bloom, xue-etal-2021-mt5, li2023bactrianx, wei2023polylm}, our analyses show that without further task-specific fine-tuning, these models are not suitable for carrying out high-quality text-editing tasks (\autoref{sec:text-editing-quality--baselines}). In the context of text editing tasks, multiple previous works have developed high-quality, general-purpose LLMs on non-English languages, restricting themselves, however, on either specific tasks \cite{rothe-etal-2021-simple, ijcai2022p0606, kementchedjhieva-sogaard-2023-grammatical, ryan-etal-2023-revisiting, krishna-etal-2022-shot, lai-etal-2022-multilingual} or specific languages \cite{alhafni-etal-2023-advancements, anschutz-etal-2023-language}. Overall, the aforementioned factors have limited the availability of high-quality \textit{multilingual text editing} (MTE) models, which has limited their usability for writing assistance across multiple tasks in languages beyond English.





 
We address these gaps with \methodnosp, a multitask, multilingual extension of \coedit \cite{raheja2023coedit}. \method models can perform text editing operations for three popular tasks: Grammatical Error Correction, Paraphrasing, and Text Simplification, in multilingual and cross-lingual settings (\autoref{fig:main_examples}) across a diverse set of seven languages, spanning six different language families (\autoref{tab:language_families}).

To build \methodnosp, we fine-tune several multilingual LLMs of varying sizes on carefully curated, largely human-annotated, parallel corpora of over 200k instructional input-output pairs, using publicly available datasets (\autoref{tab:data-sources}) for different text editing tasks. We evaluate the performance of our models extensively on text editing benchmarks in both multilingual and cross-lingual settings to demonstrate their effectiveness.

\begin{table}[t]
    \small
    \centering
    \begin{tabular}{lcl}
        \toprule
        \textbf{Language} & \textbf{ISO-639-1} & \textbf{Family} \\
        \midrule
        Arabic & \texttt{ar} & Semitic \\
        Chinese &\texttt{zh} & Sino-Tibetan \\[1.5pt]
        English &\texttt{en} & \multirow{2}{*}{Germanic} \\
        German &\texttt{de} \\[1.5pt]
        Japanese &\texttt{ja} & Japonic \\
        Korean &\texttt{ko} & Koreanic \\
        Spanish& \texttt{es} & Romance \\
        \bottomrule
    \end{tabular}
    \caption{\textbf{Set of Languages.} The seven languages, along with the ISO-639-1 code and their language family, on which we train and evaluate our models.} 
    \label{tab:language_families}
\end{table}

Our contributions are as follows:
\begin{itemize}
    \vspace{-0.2cm}
    \item This work, to the best of our knowledge, is the first to investigate multi-task, multilingual text editing via instruction tuning.
    \vspace{-0.2cm}
    \item Our models achieve strong performance on multiple text editing tasks across numerous languages and are publicly released for fostering further MTE research.
    \vspace{-0.2cm}
    \item Through a comprehensive set of controlled experiments, we provide insights on how model performance on multilingual text editing tasks is affected by various choices like model architecture, model scale, and training data mixtures. 
\end{itemize}

\begin{figure}[t]
\centering
    \includegraphics[width=0.45\textwidth, trim={1.2cm 0.4cm 3cm 2.2cm},clip]{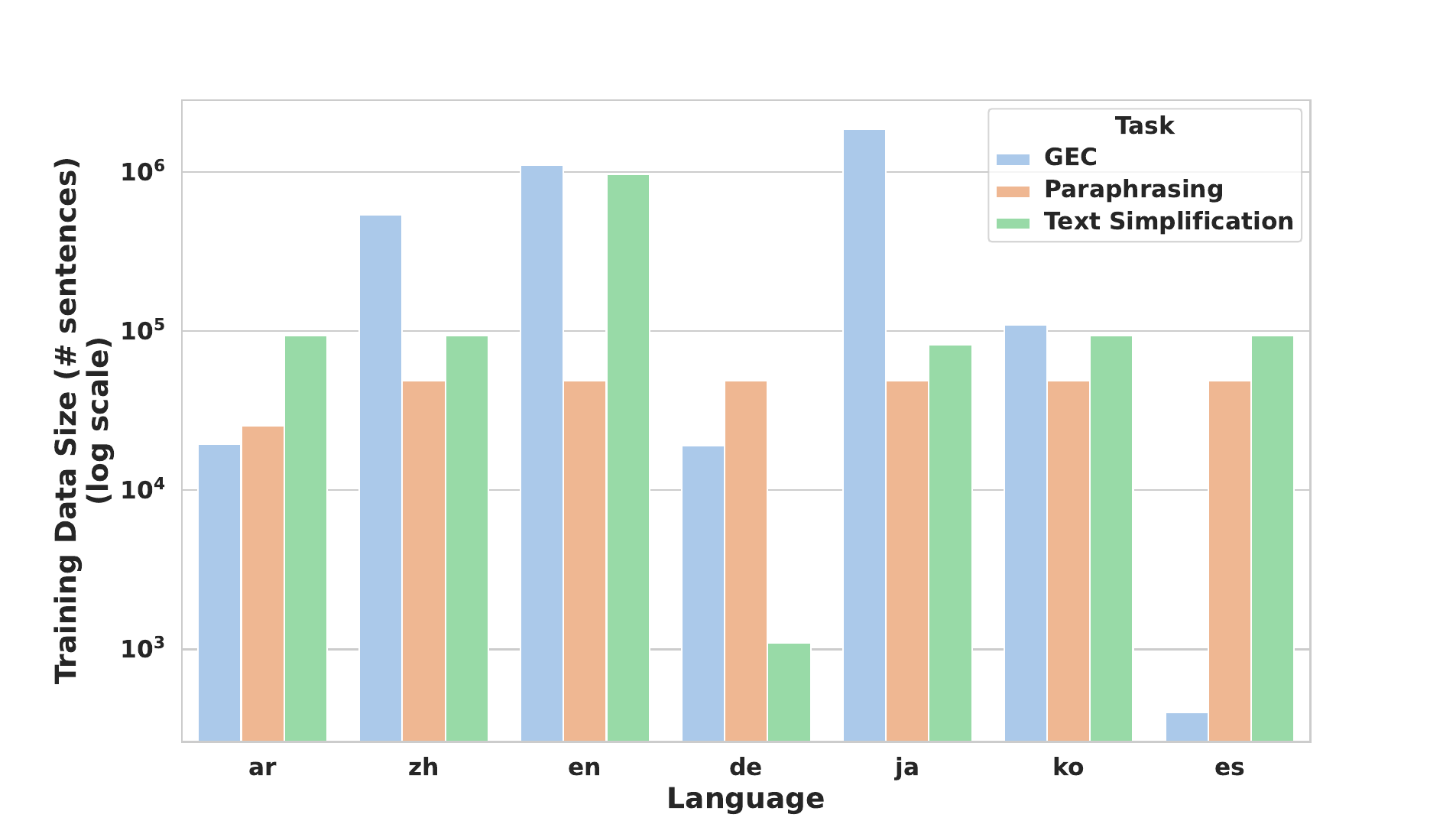}
	\caption{\textbf{Data distribution for each of the three tasks and seven languages on which we train.} The amount of data is shown in a log scale to aid visualization. }
\label{fig:data_quant}
\end{figure}

\section{Related Work}

\paragraph{Multi-lingual LLMs for Text Editing}
There is an extensive body of prior literature that has leveraged LLMs for various multi-lingual text editing tasks. These works have proposed models for text editing tasks like GEC~\cite{rothe-etal-2021-simple, ijcai2022p0606}, paraphrasing ~\cite{chowdhury2022novelty}, formality style transfer \cite{briakou-etal-2021-ola}, and text simplification ~\cite{mallinson-etal-2020-zero, martin-etal-2022-muss, ryan-etal-2023-revisiting}. However, all of these prior approaches have proposed task-specific multi-lingual models. In contrast, we propose a single unified text-editing model for all the considered tasks by leveraging the power of instruction-tuning and task-specific fine-tuning, which enables our multi-lingual models to generalize to multiple text-editing tasks.

\paragraph{Multi-lingual Instruction-Tuning}

While numerous multi-lingual instruction fine-tuned models like \citet{muennighoff-etal-2023-crosslingual, wei2023polylm} and \citet{li2023bactrianx} have been developed, they are not focused or tailored for text editing tasks, which we address by task-specific fine-tuning. 
Specific to text editing, many prior works have explored instruction tuning capable of performing multiple text editing tasks with a single model, such as GEC, simplification, sentence fusion, style transfer, and paraphrasing, to name a few \cite{mallinson-etal-2022-edit5, du-etal-2022-understanding, kim-etal-2022-improving, schick2023peer, raheja2023coedit}. However, while they are able to support multi-task text editing, they are generally mono-lingual and typically restricted to specific languages (predominantly English).
Thus, our work addresses this gap by proposing multilingual, instruction-tuned models for multiple text editing and revision tasks. 


\section{\method}

\subsection{Tasks and Languages}
We chose a broad set of languages to ensure coverage and chose text editing tasks that had multilingual, publicly available human-annotated datasets to ensure high data quality. Another criteria was to choose languages at the intersection of the publicly available corpora we could find across a large set of languages for all the tasks we considered. We refer to this as the \method dataset.

\autoref{tab:language_families} describes the languages covered in our work, whereas \autoref{fig:data_quant} depicts the amounts of training datasets that were available for all tasks and languages we considered. \autoref{app:training_testing_datasets} details all the training and testing datasets. 

\subsection{Models}
We fine-tune different versions of pre-trained multilingual LLMs (both encoder-decoder/sequence-to-sequence (Seq2Seq) and decoder-only/causal language models (CLM)) on the \method dataset using cross-entropy loss. The details of the \method models are described in \autoref{sec:medit_models}, whereas the training details are summarized in \autoref{sec:training_details}.

\section{Experiments}

\subsection{No-Edits Baseline} We first evaluate a no-edits baseline, where the output is simply a copy of the source input without the instruction. This strategy performs reasonably well on tasks where the target output largely overlaps with the input (e.g., GEC).

\subsection{Multilingual LLMs}
\label{sec:medit_models}

    \paragraph{mT5} \cite{xue-etal-2021-mt5} is a multilingual variant of T5 \cite{2020t5}, trained on the mC4 dataset, a multilingual variant of the C4 dataset extended to 101 languages. We experiment with three variants of mT5 -- \textsc{large} (770M), \textsc{xl} (3B), and \textsc{xxl} (13B) parameters. 
    \paragraph{mT0} \cite{muennighoff-etal-2023-crosslingual} is a family of multilingual Seq2Seq models capable of zero-shot following human instructions in dozens of languages. We use the mt0-\textsc{large} (1.2B), mt0-\textsc{xl} (3.7B),	and mt0-\textsc{xxl} (13B) models. These models are constructed by fine-tuning  mT5 models on the xP3 cross-lingual task mixture dataset, which consists of multilingual datasets with English prompts. As a result, mT0 models are better suited for following English prompts. We also use the mt0-\textsc{xxl-mt} variant, which is fine-tuned on the xP3mt dataset and is better suited for prompting in non-English.
    \paragraph{BLOOMZ} \cite{muennighoff-etal-2023-crosslingual} is a family of multilingual Causal Language Models (CLMs) constructed by fine-tuning BLOOM \cite{workshop2023bloom} on the xP3 dataset. We use BLOOMZ-3b and BLOOMZ-7b1 models for our experiments.
    \paragraph{PolyLM} \cite{wei2023polylm} is a set of multilingual LLMs trained on 640B tokens. We experiment with the \texttt{PolyLM-MultiAlpaca-13B} model, which is PolyLM-13B model fine-tuned on the \texttt{MultiAlpaca} dataset, consisting of 132k samples of multilingual instructions.
    \paragraph{Bactrian-X} \cite{li2023bactrianx} is a collection of lightweight adapters for LLaMA (7B and 13B) \cite{touvron2023llama} and BLOOM (7B) \cite{workshop2023bloom} on the Bactrian-X dataset, which is a multilingual parallel dataset comprising 3.4 million instruction–response pairs across 52 languages. For simplicity, we only compare against its higher-performant LLaMA-adapted versions.


\subsubsection{Large-Pretrained Decoder-only Models}
We also conduct zero-shot evaluations against state-of-the-art decoder-only LLMs that have shown impressive multilingual capabilities on a variety of NLP tasks leveraging the power of in-context learning \cite{lai2023chatgpt, openai2023gpt4}.
 
\paragraph{GPT3.5} (also referred to as ChatGPT),\footnote{\url{https://openai.com/blog/chatgpt}} is an improved version of GPT3 \cite{gpt3} optimized for chat. We use the \texttt{gpt-3.5-turbo0613} model from the OpenAI API.\footnote{\url{https://api.openai.com}}
\paragraph{GPT4} \cite{openai2023gpt4} is the latest iteration of the GPT models and is also optimized for chat. We use the \texttt{gpt-4-0613} model from the OpenAI API.

While we recognize that these models may not be explicitly optimized or trained for multi-lingual settings, considering that they have been trained on massive amounts of web-scale data, these models have been shown to have multi-lingual capabilities \cite{lai2023chatgpt}, hence, we consider them as one of our baseline groups.

\subsection{Training Setup}
\label{sec:training_details}
We perform instruction tuning for all our models by crafting custom prompts for each of the 21 task-language combinations (seven languages, three tasks). Similar to \coeditnosp, for each task-language combination, depending on the number of ways the instructions can be translated without altering the meaning, we write between 14 and 27 instructions by automatically translating each one from English and verifying the accuracy of the translations by asking native language speakers to evaluate and correct them. The total number of task-language instructions is 365, which can be found in \autoref{app:verbalizers}. 
We explore three different multilingual and cross-lingual instructional settings, depending on the language of the prompt, where the editing instruction could be in (a) \textbf{English}, (b) the same language as the text being edited (\textbf{Native}), and (c) a random language which may or may not be the same as the language of the text being edited (\textbf{Random}). With this definition, English and Random are cross-lingual text editing tasks, and Native is a multilingual text editing task.
In all settings, the input-output pairs are in the same language, but only the language of the instruction changes. 

We train all models on 8xA100 80G GPU instances for five epochs. For the PolyLM and Bactrian-X models (>7B parameters), we also use LoRA \citep{hu2022lora} to lower the number of trainable parameters and increase the batch size. 


\subsection{Evaluation}
\label{sec:evaluation}
For GEC evaluation, we follow prior work on each language we report on and use the appropriate GEC metric accordingly. Mainly,  we use the MaxMatch (M\textsuperscript{2}) Scorer \cite{dahlmeier-ng-2012-better}, ERRANT \cite{bryant-etal-2017-automatic}, and GLEU \cite{napoles-etal-2015-ground,napoles2016gleu} as our evaluation metrics. The M\textsuperscript{2} Scorer and ERRANT compare the edits made by a GEC system against annotated reference edits and calculate the precision (P), recall (R), and F\textsubscript{0.5} (i.e., weighing precision twice as much as recall). GLEU computes the precision of the n-grams that overlap with the references but not the original texts and penalizes n-grams that overlap with the original texts but not the references.

For simplification, we follow \citet{ryan-etal-2023-revisiting} and use SARI \cite{xu-etal-2016-optimizing} and BLEU \cite{papineni2002bleu} for evaluation. SARI is the average of the F1 score for adding, keeping, and deleting n-grams $(n \in {1, 2, 3, 4})$ and has been shown to correlate with human judgments of simplicity \cite{xu-etal-2016-optimizing}. BLEU, on the other hand, is a common metric in machine translation and is used as a check for grammatical and meaning preservation. We compute all metrics using the EASSE evaluation suite \cite{alva-manchego-etal-2019-easse}.

We evaluate paraphrasing on two criteria and metrics: diversity and semantic similarity. For diversity, we use Self-BLEU \cite{10.1145/3209978.3210080} to measure the diversity of the paraphrases relative to the given source and reference texts. We use Semantic Similarity to measure meaning preservation. Specifically, we use \textit{m}USE \cite{yang-etal-2020-multilingual} for this, as it is the best-performing multilingual sentence similarity model that supports all the languages in our work. We also considered other notable works that have made significant progress on multilingual sentence similarity, such as Multilingual-SBERT \cite{reimers-gurevych-2020-making} and LaBSE \cite{feng-etal-2022-language}. However, we found them unsuitable for our purposes as they were either limited by the languages they support or suffered from lower performance for multilingual sentence similarity.


\section{Quantitative Results: MTE Quality}
\label{sec:text_editing_quality}

\vspace{-0.2cm}
We split the models into three main groups. The first group (a) consists of the ``no edits'' baseline, the second group (b) is the untrained baseline, where models are evaluated in a zero-shot setting without any task-specific fine-tuning, while the third group (c) is our set of multi-lingual models trained on task-specific datasets. For all our experiments, we aggregate the models' performance by text editing tasks. Specifically, we aggregate the metrics for each task using the harmonic means of its constituents. Specifically, we use (1-Self-BLEU)\footnote{We subtract Self-BLEU from 1 because lower is better in terms of making changes to the source text.} and Semantic Accuracy for Paraphrasing, SARI and BLEU for Simplification, and $F_{0.5}$ and GLEU for GEC. We scale all metrics to lie between 0 and 100. We show full results on all models in \autoref{app:main_results} for the best-performing setup.

\begin{figure}[t]
    \centering
    \includegraphics[width=0.48\textwidth, trim={0.2cm 0.2cm 0.2cm 0.3cm}]{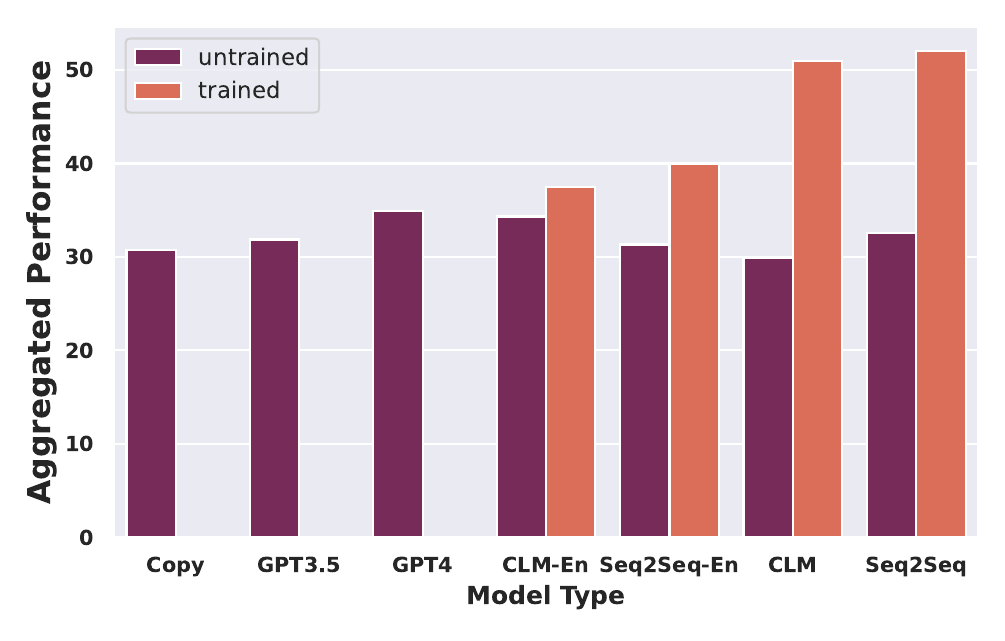}
    \caption{\textbf{Overall performance comparison of all baselines against trained models.} We calculate the aggregated performance across all tasks using the harmonic mean of task-specific scores. Baselines are ``No Edits (Copy)'', ``English-only'' (``-En,''), and our trained models are marked as ``CLM'' and ``Seq2Seq,'' respectively. The aggregated performance is calculated as described in \autoref{sec:text_editing_quality}.}
    \label{fig:baselines}
    \vspace{-0.4cm}
\end{figure}

\subsection{Baselines}
\label{sec:text-editing-quality--baselines}
In \autoref{fig:baselines}, we report the results of our trained models against various baselines by aggregating the performance on all tasks (as detailed in \autoref{sec:text_editing_quality}).
\vspace{-0.1cm}

\paragraph{No Edits (Copy) Baseline}
We observe that not making any edits leads to a performance that is on par with the untrained versions of all models, which highlights the limitations of the n-gram overlap-based metrics. 
\vspace{-0.1cm}

\paragraph{Untrained Baseline}
Similar to \citet{raheja2023coedit}, a core contribution of this work is to push the performance of small- ($\sim$1B parameters) to medium-sized (1-15B parameters) LLMs for common text editing tasks across multiple languages. This drives the need for fine-tuning task-specific and language-specific datasets. For this work, we compare our fine-tuned models against their non-fine-tuned counterparts. We find a substantial gap between the untrained models and their trained counterparts, highlighting the impact of task- and language-specific fine-tuning for the tasks.
\vspace{-0.1cm}

\paragraph{English-Only Baseline}
In this experiment, we analyze the ability of multilingual LLMs to adapt to different text editing tasks across different languages by fine-tuning them in the most prominent high-resource language (English). We fine-tune all the multi-lingual models on just the English subsets of the training data, as it is the largest in terms of quality and quantity. This experiment tests the need for language-specific data. Similar to the previous result, we observe that the gap between the untrained versions of English-only models is relatively small vs. the ones trained on the full dataset; it increases significantly with language-specific fine-tuning.
\vspace{-0.1cm}

\paragraph{State-of-the-art LLMs}
Additionally, we also evaluate against the most powerful commercially available LLMs on their ability to perform MTE. Specifically, we evaluate GPT3.5 and GPT-4 in a zero-shot setting. Although these models have been shown to exhibit strong zero-shot performance on a variety of NLP tasks, we find that the overall performance of both models is close to most untrained baselines. This can be attributed to the rather limited multilingual capabilities of GPT3.5, which often lead to outputs being generated in other languages (English in particular). To some extent, the verbosity of responses is highly detrimental to the performance, especially for GPT4 as it gets penalized by the automatic metrics (especially for GEC).

The rest of this section analyzes different aspects of the quantitative performance of our models.

\begin{figure}[!t]
    \centering
    \includegraphics[width=0.48\textwidth, trim={0.2cm 0.2cm 0.2cm 0.3cm}]{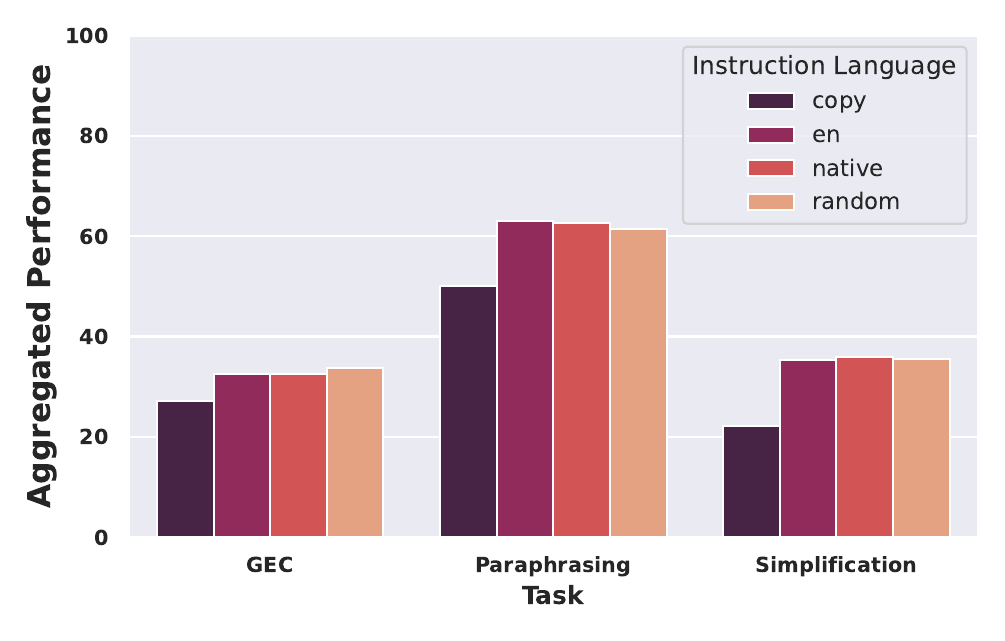}
    \caption{\textbf{Aggregated performance on different tasks broken down by instruction language}. Apart from some minor fluctuation, there is no significant impact of instruction language on our results.}
    \label{fig:prompt_lang}
\end{figure}

\begin{figure*}
    \centering
    {\includegraphics[width=0.32\textwidth, trim={0.32cm 0.2cm 0.25cm 0.1cm},clip]{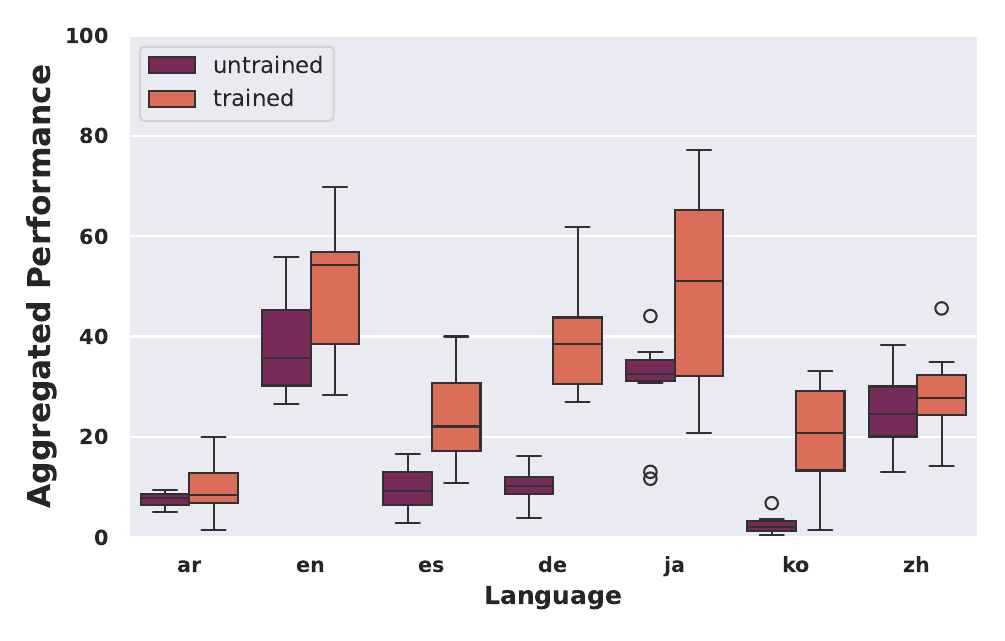}}
    {\includegraphics[width=0.32\textwidth, trim={0.25cm 0.2cm 0.32cm 0.1cm},clip]{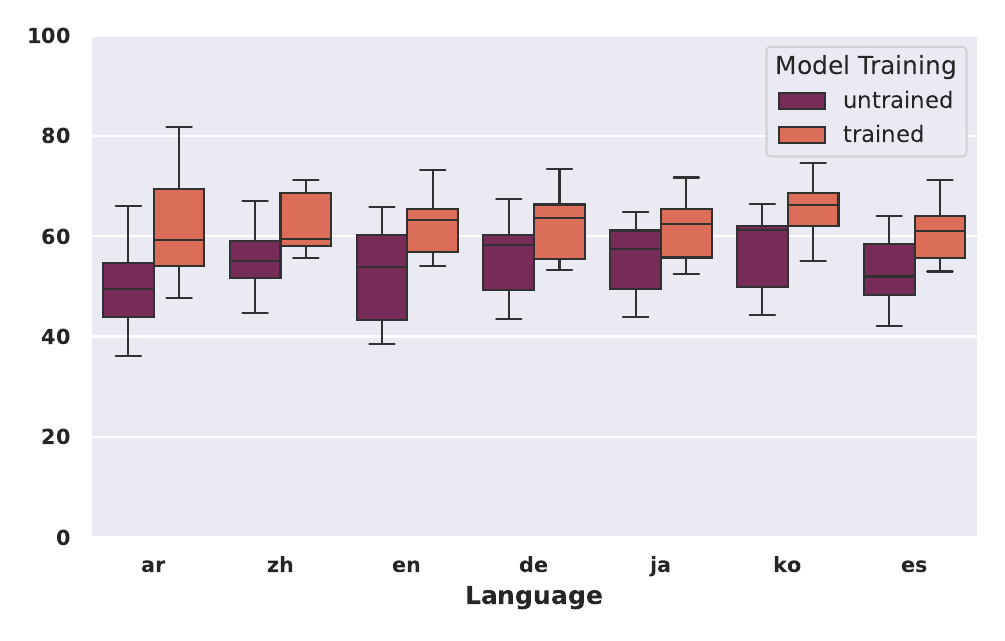}} 
    {\includegraphics[width=0.32\textwidth, trim={0.25cm 0.2cm 0.32cm 0.3cm},clip]{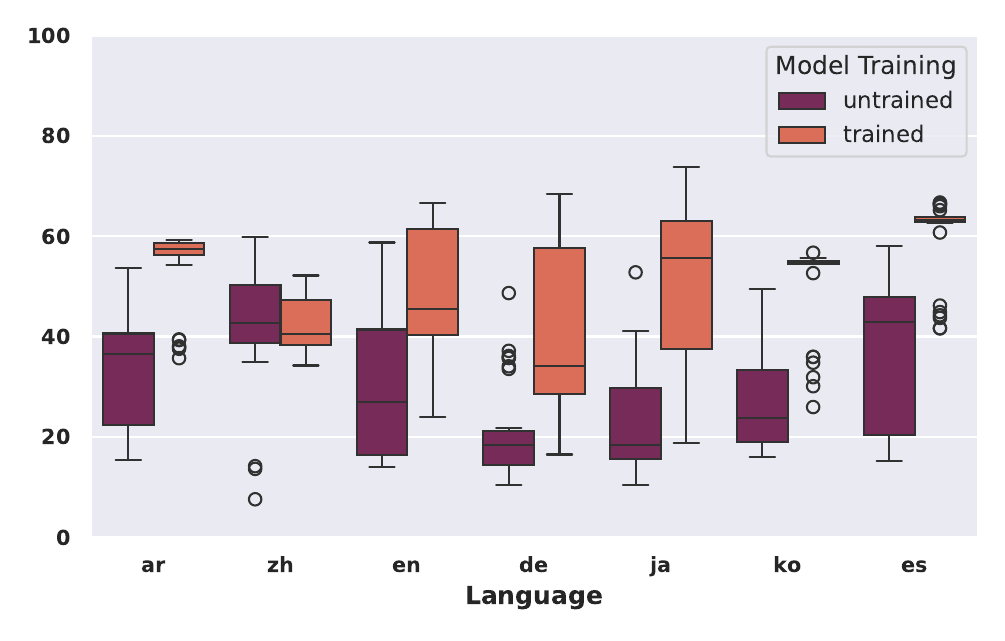}}
    \caption{\textbf{Aggregated model performance by language} (for GEC, Paraphrasing, and Simplification). For each task, we aggregate the relevant metrics as described in \autoref{sec:text_editing_quality} and split them by model training.}
    \label{fig:lang_tasks}
\end{figure*}

\subsection{Model Performance by Language}
\label{sec:language}
In this section, we analyze the performance of different MTE models by language (\autoref{fig:lang_tasks}). It is interesting to note that Paraphrasing exhibits a rather steady performance across languages. This can partially be attributed to the weakness of the evaluation metrics as they rely mostly on n-gram overlap, on the multilingual pre-training of the LLMs where they are exposed to medium-large corpora of nearly all the languages, but also that with the increase in the model size and fine-tuning, they tend to make fewer changes, thus, leading to higher scores. For simplification, the variance in the performance across languages can be attributed to the amount and quality of training data available for each task. For instance, for German, only 1.1k training data points were available, which leads to models not only showing a great improvement in performance with fine tuning, but also a great variance. Similarly, the training data is very noisy for Japanese, which leads to a similar effect. For GEC, we observe that the performance varies a lot by language, indicating the challenging nature of the task. This can be partially attributed to the frequency and the type of errors in each dataset, a phenomenon we see in Arabic datasets. For instance, Arabic datasets contain subtle and frequent errors made by native/L1 speakers, whereas the data for all other languages consists of sparse and infrequency errors made by L2 learners, which could explain the abnormally low quality (on Arabic GEC) of our best-performing model.
Moreover, the quality of the training GEC data available also leads to varying performance across languages.
\vspace{-0.1cm}

\subsection{Language of Instruction}
\vspace{-0.1cm}
Here, we analyze the effect of the language of the instruction used to instruct the model. As mentioned in \autoref{sec:training_details}, we have three configurations for this set of experiments:
\vspace{-0.1cm}

\paragraph{English-language Instructions} We train the first set of multilingual \method models with just English instructions. These are MTE models capable of performing cross-lingual text editing, trained on data where the instruction is always in English.
\vspace{-0.1cm}
\paragraph{Native-language Instructions} We train the next set of multi-lingual \method models with instructions in their native language. These models are capable of performing MTE, where the language of edit instructions is the same as the language of texts being edited.
\vspace{-0.1cm}
\paragraph{Randomized-language Instructions} Finally, we also explore the cross-lingual text editing (\autoref{fig:main_examples}) abilities of multi-lingual LLMs. To do so, we modify our dataset by appending an edit instruction from a randomly chosen language (different from the language of the edited source-target text pair) and train our models on this cross-lingual-prompted dataset. 

\autoref{fig:prompt_lang} shows the effect of instruction language on performance on all three tasks. We note that there is no significant difference in performance between the different settings. This is likely because in each setting, owing to its multilingual instructional pre-training, the model is able to adapt well to the language of the instruction in the fine-tuning phase, hence focusing mostly on the specific tasks.

\begin{figure*}
    \centering
    {\includegraphics[width=0.32\textwidth, trim={0.2cm 0.2cm 0.2cm 0.3cm},clip]{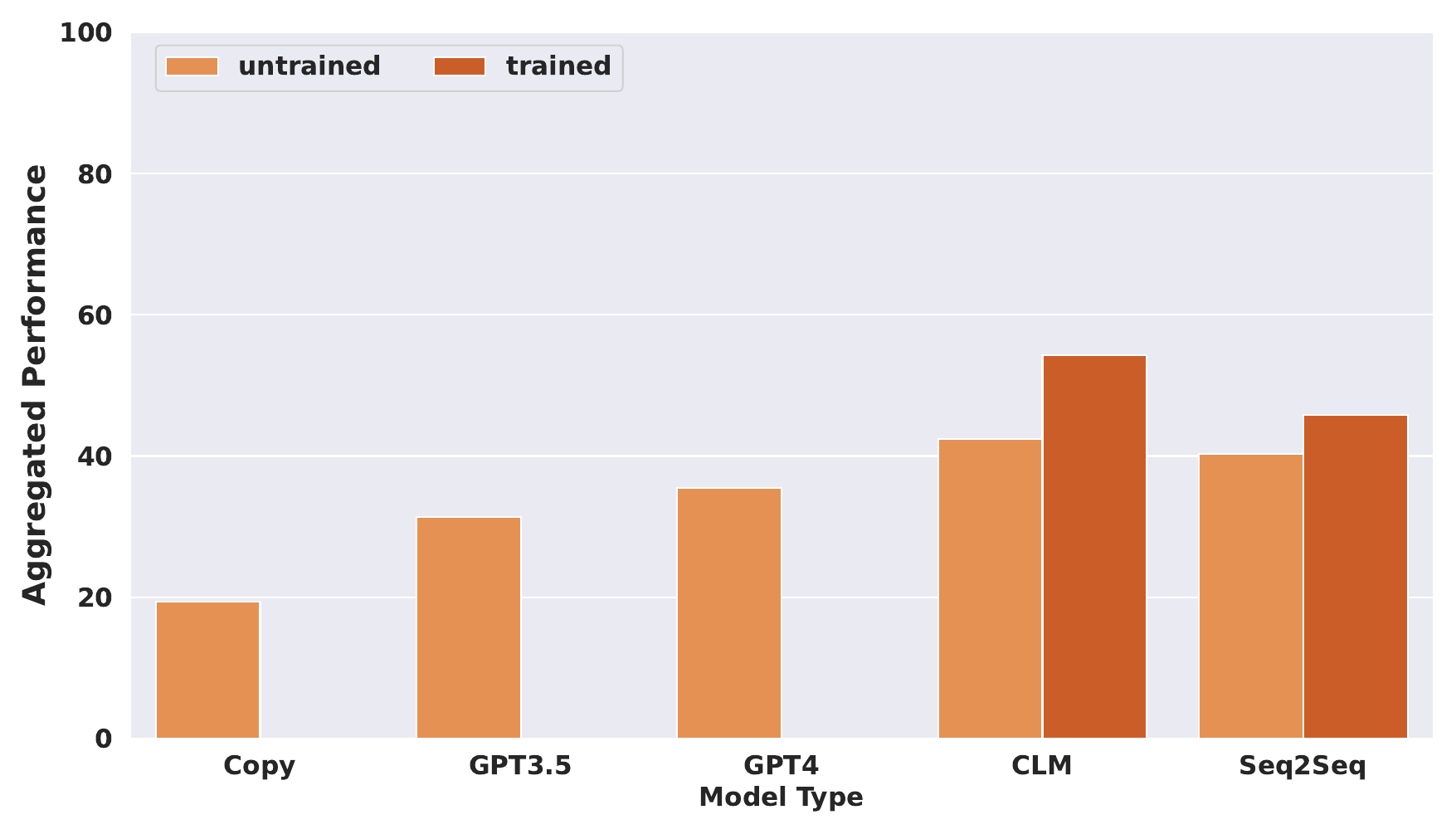}}
    {\includegraphics[width=0.31\textwidth, trim={1.1cm 0.2cm 0.2cm 0.3cm},clip]{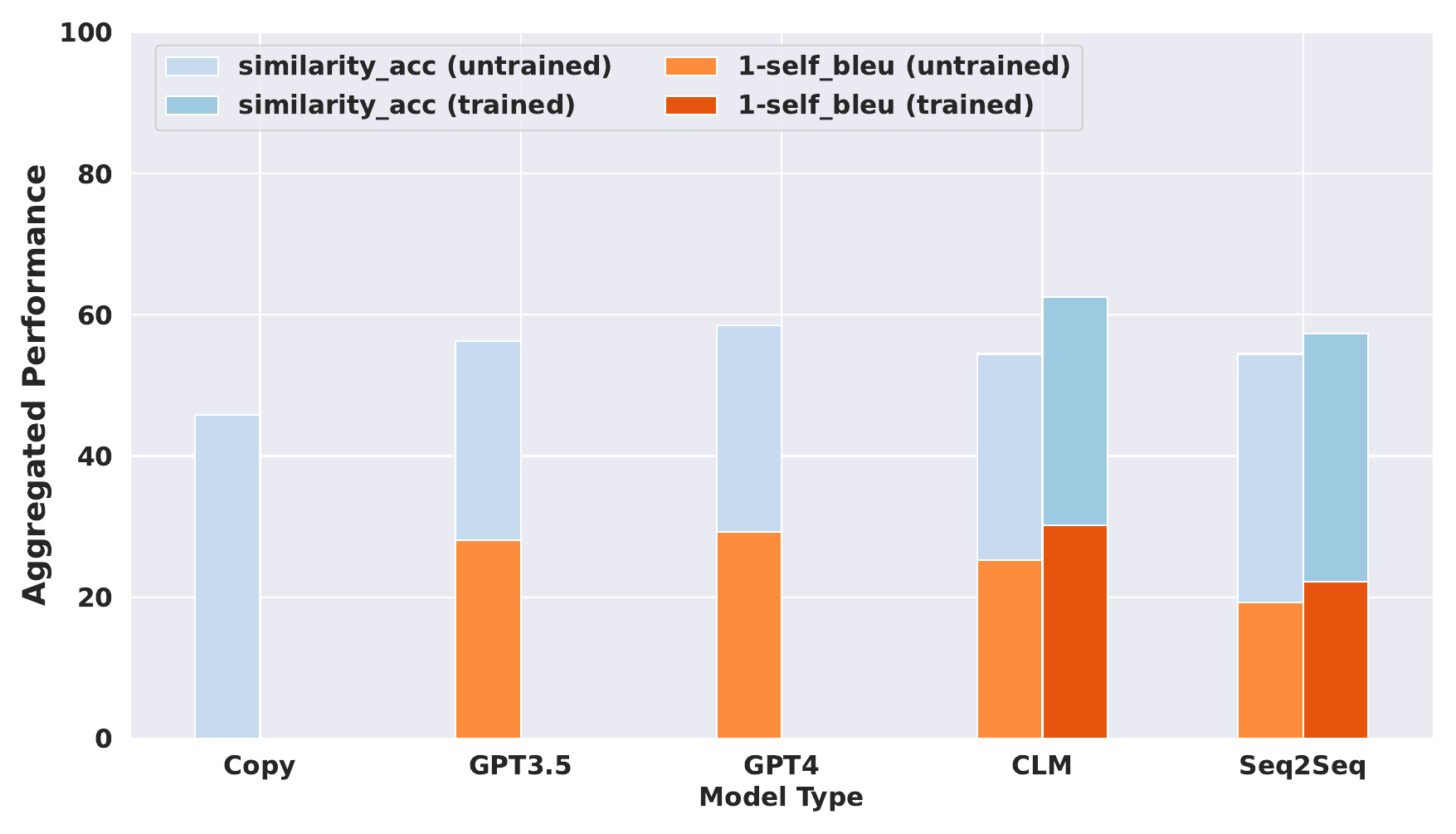}}
    {\includegraphics[width=0.31\textwidth, trim={1.1cm 0.2cm 0.2cm 0.3cm},clip]{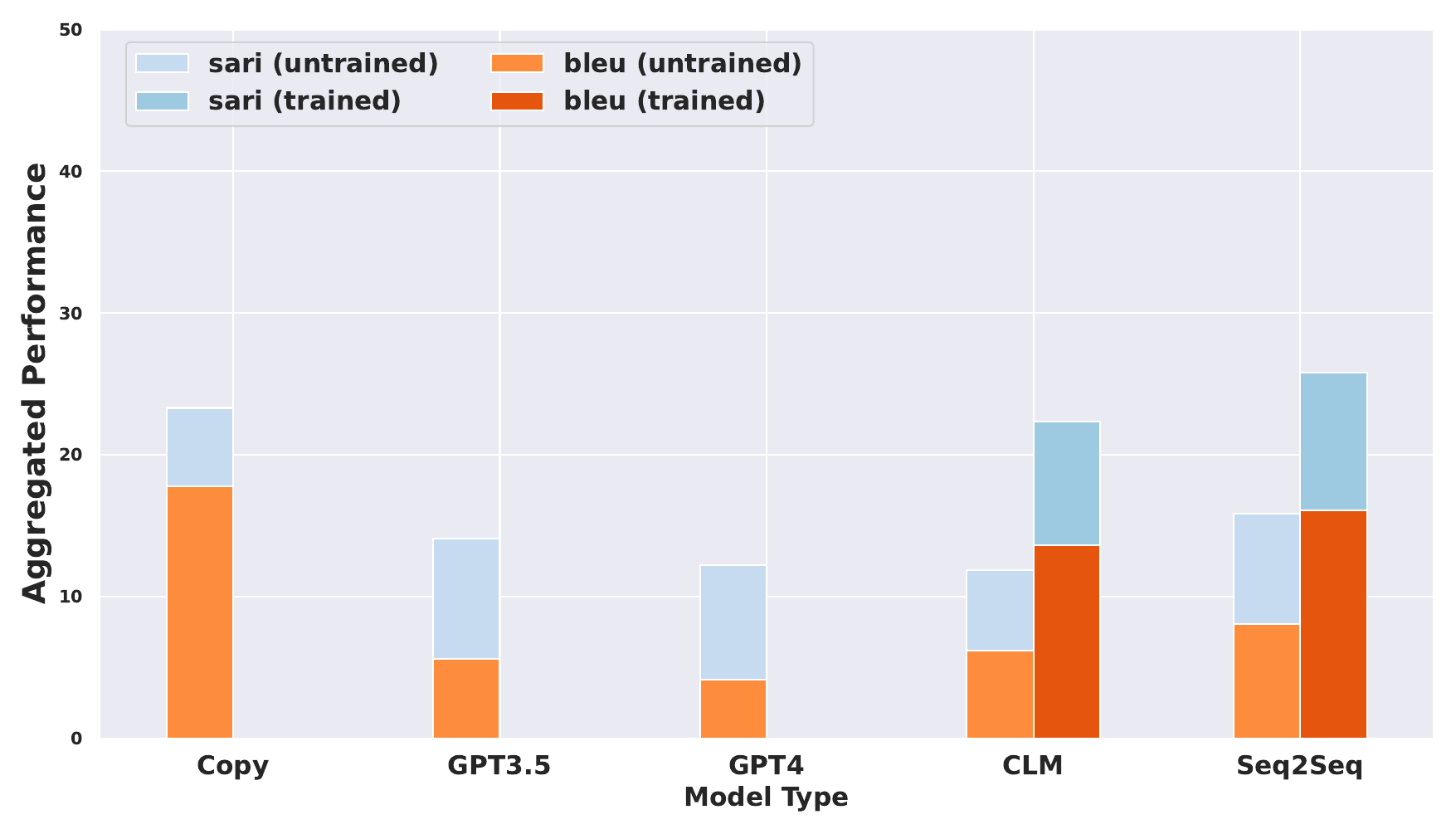}}    
    \caption{\textbf{Aggregated performance by model type} (for GEC, Paraphrasing, and Simplification). For each task, we aggregate the relevant metrics as described in \autoref{sec:evaluation} and split them by model type (CLM vs Seq2Seq), including the copy baseline.}
    \label{fig:model_type}
\end{figure*}

\subsection{Effect of Model Architecture}
We present the task-specific results across each model type in \autoref{fig:model_type}, observing that CLMs generally are either on par or outperform the rest of the models with GPT3.5, yielding the lowest results.

BLEU relies on n-gram overlap artificially boosting the scores, which highlights the disadvantage of the Copy baseline. Also, GPT3.5 and GPT4 consistently perform poorly in comparison to the rest of the models, which is especially surprising given GPT4's multilingual capability. In addition, this could be an artifact of the metrics since RLHF can produce excellent results that have little overlap with the references, and future human studies may shed more light on the issue.

CLMs and Seq2Seq models perform similarly on GEC and Paraphrasing, while Seq2Seq performs better on simplification. We posit that this discrepancy happens due to shorter generations from the Seq2Seq models since we observe that the seq2seq models generate significantly shorter sequences than the expected distribution ($D = 0.06, p < 0.001$),\footnote{Using a two-tailed Kolmogorov-Smirnov test, compared against the distribution of lengths of the reference texts.} increasing the BLEU scores, which are sensitive to the prediction length, whereas CLMs tend to generate longer sequences ($D = 0.27, p < 0.001$). Looking at the SARI scores, we observe that the two model types do not differ significantly ($p~>~.05$), indicating similar performance overall.


\subsection{Effect of Model Scale}
\begin{figure}[t]
    \centering
    \includegraphics[width=0.48\textwidth, trim={0.2cm 0.2cm 0.2cm 0.3cm}]{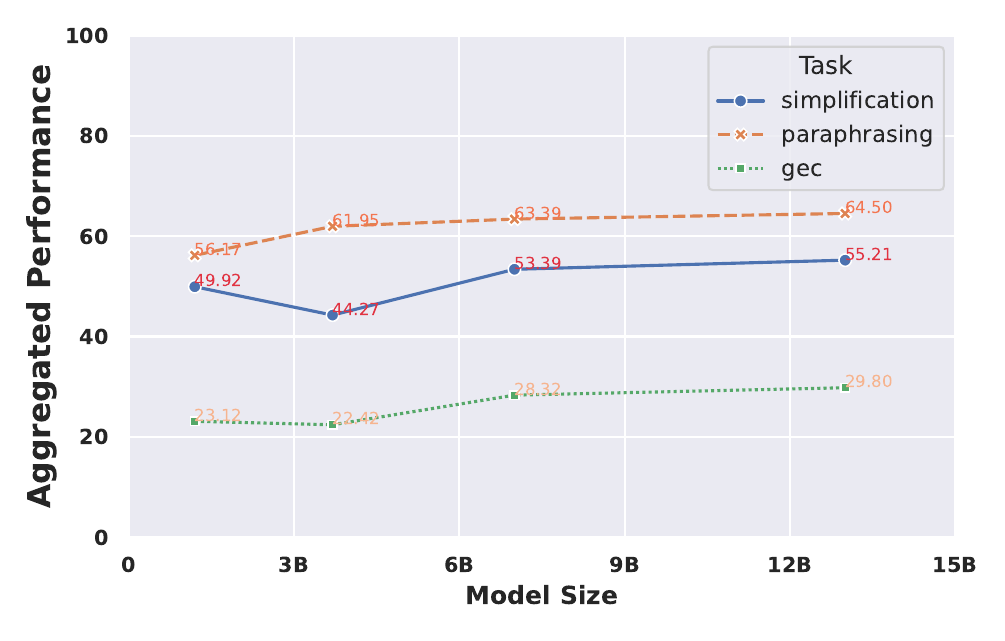}
    \caption{\textbf{Aggregated model performance on different tasks broken down by parameter size.} For visualization reasons, we group the 1.2B and 1.7B models and the 7B and 7.1B models together.}
    \label{fig:model_scale_tasks_params}
\end{figure}

In \autoref{fig:model_scale_tasks_params}, we describe the overall performance of MTE models by size aggregated over the three tasks. It is evident that scaling model size generally increases overall performance significantly, thus reinforcing the effectiveness of model scaling. We also note that all three tasks display similar trends, with relative improvements being the greatest in GEC (\textbf{28.8\%}) and Paraphrasing (\textbf{14.8\%}).

\subsection{Effect of Task-specific Data}
\begin{figure*}[t]
    \centering
    {\includegraphics[width=0.34\textwidth, trim={0.2cm 0.2cm 0.2cm 0.3cm},clip]{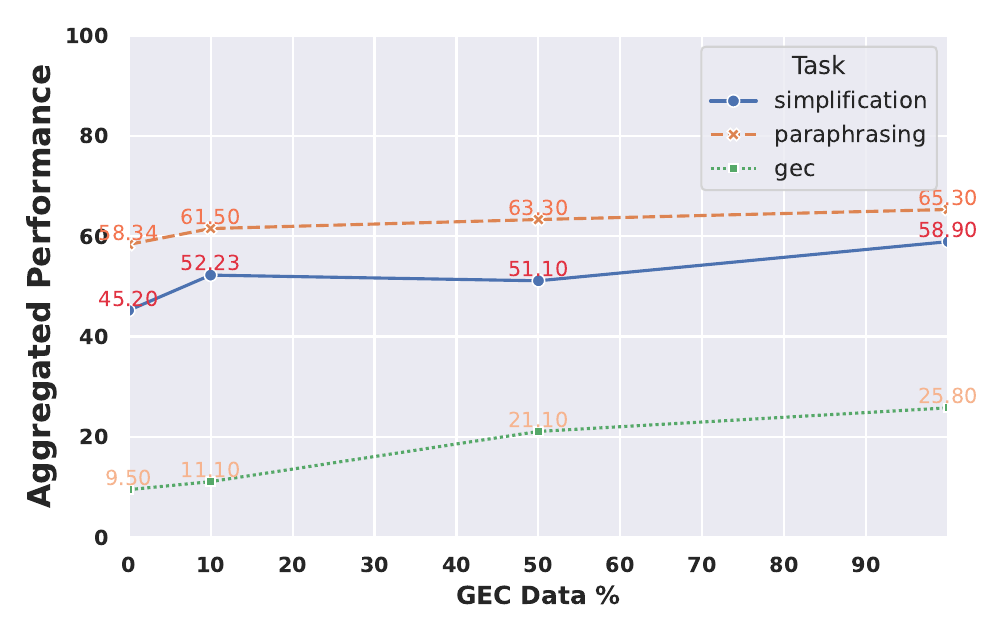}}    
    {\includegraphics[width=0.32\textwidth, trim={1.1cm 0.2cm 0.2cm 0.3cm},clip]{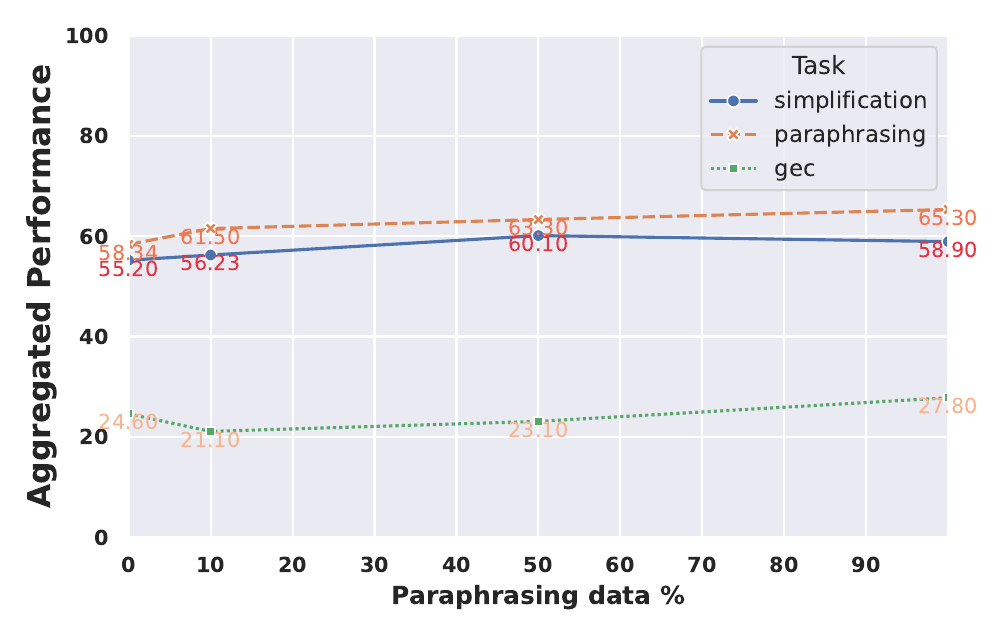}}    
    {\includegraphics[width=0.32\textwidth, trim={1.1cm 0.2cm 0.2cm 0.3cm},clip]{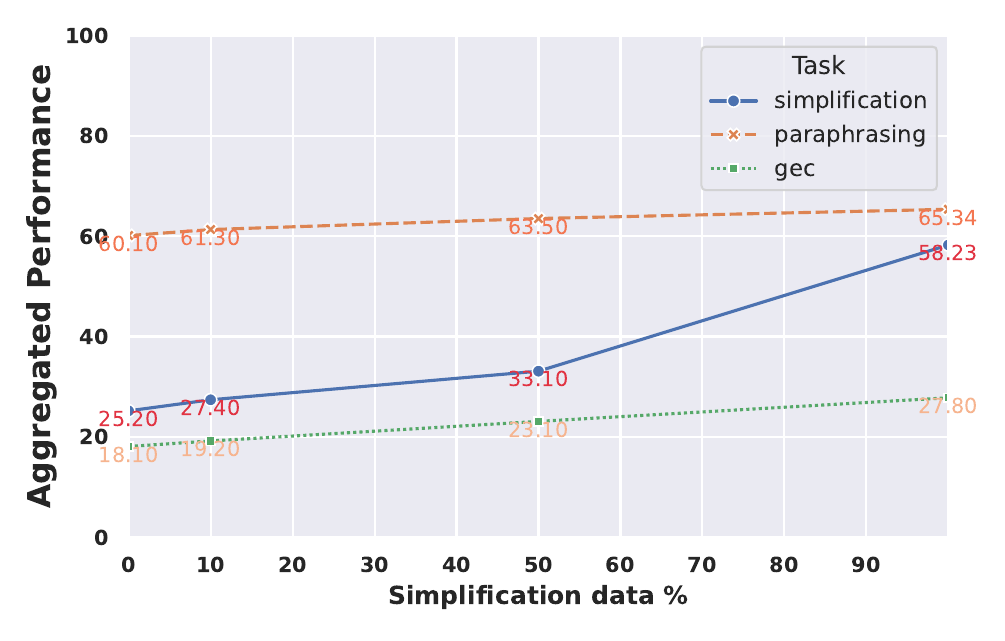}}
    \caption{\textbf{Aggregated model performance by varying amounts of data samples}: (0\% to 100\%) by task (in the order: GEC, Paraphrasing, Simplification). We aggregate the scores as described in \autoref{sec:text_editing_quality}.}
    \label{fig:model_type_tasks}
    \vspace{-0.1cm}
\end{figure*}

To understand the effect of task-specific data on model performance, we systematically ablate the proportion of training data for each task. Specifically, we conduct three groups by varying the amounts of training data across a given task between 0\%, 10\%, 50\%, and 100\% while keeping the amount of training data across other tasks at 100\%, the results of which are shown in   \autoref{fig:model_type_tasks}. We observe that: (a) Performance on the ablated task generally improves as the amount of training data for that task increases, as expected. As the proportion of training data increases, so does the performance on the specific task. (b) We also note a synergistic relationship between some tasks where training data from one task helps improve performance on a different task. For example, as we add training examples from GEC, we also notice an improvement in model performance on simplification ($58.90$ vs $45.20$) and paraphrasing ($65.30$ vs $58.34$). Similar trends also hold when we add data for the other two tasks. We believe our model (which is inherently multi-task) enables us to leverage such synergy between text editing tasks better, as compared to task-specific models.


\begin{table}[!ht]
\vspace{-0.1cm}
\centering
\tiny
\begin{tabular}{p{0.06\linewidth}p{0.1\linewidth}p{0.13\linewidth}@{\hskip 1mm}rlp{0.06\linewidth}l}
\toprule
\textbf{Task} & \textbf{Language} & \textbf{Dataset} & \textbf{Test Size} & \textbf{Metric} & \textbf{\method} & \textbf{SOTA} \\
\midrule
\parbox[t]{2mm}{\multirow{3.5}{*}{\rotatebox[origin=c]{90}{GEC}}}  & Romanian &  RoGEC  & 1518 & GLEU & 45.58 & -- \\
\cmidrule{2-7}
    & \multirow{2}{*}{Hindi} &  \multirow{2}{*}{HiWikEd} & \multirow{2}{*}{13187} & ERRANT & 32.61 & 49.4 \\
    &  &   &  & GLEU & 68.91 & 80 \\
\cmidrule{2-7}
\parbox[t]{2mm}{\multirow{4.5}{*}{\rotatebox[origin=c]{90}{Simplification}}}  & \multirow{2}{*}{Italian} & Simpitiki  & 176 & SARI & 47.84 & 24.27 \\
                                &  & PaCSSS-IT & 63007 & BLEU & 41.11 & 36.92 \\
\cmidrule{2-7}
    & \multirow{2}{*}{Hindi} & \multirow{2}{*}{IndicSS} &  \multirow{2}{*}{42771} & SARI & 40.08 & -- \\
    & &   & & Rouge-L & 19.92 & 45.57 \\
\cmidrule{2-7}
\parbox[t]{2mm}{\multirow{5.5}{*}{\rotatebox[origin=c]{90}{Paraphrasing}}}  & \multirow{2}{*}{French} & \multirow{2}{*}{PAWS-X } & \multirow{2}{*}{903} & Self-BLEU & 69.06 & -- \\
& & & & SA & 98.38 & -- \\  
\cmidrule{2-7}
    & \multirow{3}{*}{Hindi} & \multirow{3}{*}{IndicPara. \hfill\break} & \multirow{3}{*}{10000} & Self-BLEU & 23.91 & -- \\
    && & & SA & 92.06 & -- \\
    &&  & & iBLEU & 14.2 & 18.55 \\
\bottomrule
\end{tabular}
\caption{\textbf{Zero-shot evaluation results on the language generalization experiments.} We present the scores achieved by our best-performing model (\textbf{Our score}) along with the current \textbf{SOTA} results\protect\footnotemark. Wherever possible, we report the metrics reported in the SOTA papers, and if not available, we report commonly used ones by the literature. Note that we focus only on languages that the LLMs have seen during pre-training \autoref{sec:generalization-to-new-languages}.}


\label{tab:data-new-langs}
\end{table}

\footnotetext{Test datasets for new languages include: RO-GEC: \citep{cotet2020rogec}, HI-GEC: \citep{sonawane-etal-2020-generating}, IT-SIMP: \citep{tonelli2016simpitiki,brunato-etal-2016-paccss}, HI-SIMP: \citep{kumar-etal-2022-indicnlg}, FR-PARA: \citep{yang-etal-2019-paws}, HI-PARA: \citep{kumar-etal-2022-indicnlg}.}

\subsection{Generalization to new languages}
\label{sec:generalization-to-new-languages}
\vspace{-0.1cm}
We also explore the capabilities of our \method models on new languages. For every task, we chose two new languages not present in the training set: one related to the language families covered in our training dataset (\autoref{tab:language_families}) and one belonging to a language family not present in the training dataset. We only considered the languages that the underlying LLM had as part of its pre-training corpora so as to ensure the models had some understanding of the languages in question. \autoref{tab:data-new-langs} provides a summary of the languages considered and the datasets they were sourced from, as well as the results of the language generalization experiments. We follow the same metrics for all the tasks as in \autoref{sec:evaluation}.

\method models are competitive on many unseen languages as compared to the monolingual state-of-the-art, especially on \texttt{it}-Simplification, and \texttt{hi}-GEC and Paraphrasing.\footnote{Since no multilingual models have attempted to perform the considered tasks in the respective languages, we are unable to report the metrics on others. We also do not provide any comparisons if the language-specific metrics either do not exist or are reported using different metrics.}








\section{Human Evaluations}

\begin{table}[t!]
  \centering
  \tiny	
  \begin{tabular}{@{}p{0.0001\textwidth}cccccc@{}}
    \toprule
    &\textbf{Language} & \textbf{V. Good} & \textbf{Good} & \textbf{Neutral} & \textbf{Bad} & \textbf{V. Bad}\\
    \midrule
     \parbox[t]{2mm}{\multirow{7}{*}{\rotatebox[origin=c]{90}{\textsc{fluency}}}} & Arabic & 25.00 & 14.29 & 17.86 & 14.29 & 28.57 \\
    &Chinese & 56.67 & 13.33 & 23.33 & 3.33 & 3.33  \\
    &English & 56.67 & 23.33 & 13.33 & 3.33 & 3.33 \\
    &German & 30.0 & 56.67 & 3.33 & 6.67 & 3.33 \\
    &Japanese & 50.0 & 4.54 & 22.72 & 13.63 & 9.09 \\
    &Korean & 39.13 & 21.74 & 17.39 & 13.04 & 8.70 \\
    &Spanish & 63.33 & 10.00 & 13.33 & 10.00 & 3.33 \\
    \midrule
    \parbox[t]{2mm}{\multirow{7}{*}{\rotatebox[origin=c]{90}{\textsc{adequacy}}}} & Arabic & 21.43 & 14.29 & 10.71 & 25.00 & 28.57 \\
    &Chinese & 56.67 & 16.67 & 6.67 & 10 & 10.0 \\
    &English & 62.33 & 18.32 & 9.09 & 9.09 & 1.16 \\
    &German & 33.33 & 63.33 & 0.0 & 3.33 & 0.0 \\
    &Japanese & 63.63 & 4.55 & 4.55 & 18.18 & 9.09 \\
    &Korean & 41.67 & 16.67 & 12.50 & 20.83 & 8.33 \\
    &Spanish & 60.0 & 6.67 & 6.67 & 13.33 & 13.33 \\
    \midrule
    \parbox[t]{2mm}{\multirow{7}{*}{\rotatebox[origin=c]{90}{\textsc{accuracy}}}} & Arabic & 21.43 & 3.57 & 10.71 & 35.71 & 28.57 \\
    &Chinese & 3.45 & 13.79 & 17.24 & 51.72 & 13.79 \\
    &English & 37.93 & 32.23 & 8.33 & 18.18 & 3.33 \\
    &German & 30.0 & 40.0 & 23.33 & 6.67 & 0.0 \\
    &Japanese & 34.48 & 10.34 & 3.45 & 20.69 & 31.03 \\
    &Korean & 18.52 & 18.52 & 11.11 & 14.81 & 37.04 \\
    &Spanish & 37.93 & 24.14 & 6.90 & 3.45 & 27.59 \\
    \bottomrule    
  \end{tabular}
  \caption{\textbf{Results of the human evaluation of the model output across three criteria.} For each of the criteria, expert human annotators rate the system output, and we note the frequency of their rating (\%).}
  \vspace{-0.3cm}
 \label{tab:human-evals}
\end{table}

We conduct human evaluations of our model outputs by proficient linguists (and native language speakers of the respective languages) on 50 test inputs (per task- language) to ensure they meet the instructional task-specific constraints across the various languages since text editing is often subjective, and automatic metrics are often limited in their effectiveness and accuracy. We evaluate our best-performing \method model\footnote{\texttt{bactrian-x-llama-13b-merged}} based on its strong performance (\autoref{tab:main_results}). Specifically, we conduct a qualitative evaluation where each annotator is shown an instructional input and output from the model and asked to rate the quality of the output on three criteria: \textbf{fluency} (Is the output grammatically correct and sound like it was written by a native speaker of the language?), \textbf{adequacy} (does the output preserve the meaning of the input?), and \textbf{accuracy} (did the model make the desired edits according to the given edit instructions?) of the edited texts, on a Likert scale ranging from \textit{Very Bad} to \textit{Very Good}. We collect two annotations for each data point and adjudicate the conflicting
judgments with the annotators. The annotation guidelines are provided in \autoref{app:annotation_guidelines}. Table \ref{tab:human-evals} shows the results of the evaluation. The expert annotators generally rate the model outputs as \textit{Good} or \textit{Very Good} across nearly all languages. For Arabic, the preferences are more balanced across the scale and sometimes even leaning towards the \textit{Bad} or \textit{Very Bad} across all criteria, which confirms our findings on the automatic metrics as well, indicating that while our model performs very well on languages such as English, German, Chinese, and Spanish, it still has a long way to go in terms of performance for languages such as Arabic in terms of quality, and on Chinese, Japanese, Korean, and Spanish on accuracy.

\section{Conclusions}
We present \method – an open-sourced dataset and set of multilingual instruction-tuned LLMs capable of following natural language instructions in seven languages to perform various textual editing tasks. 
It is the first publicly available set of models that heavily outperforms numerous multilingual LLMs on multiple tasks in different languages. 
Positive feedback from human evaluations shows that \method can assist writers with various aspects of the text revision process at scale by following natural language instructions in multiple languages.
Experiments on various multilingual NLP tasks demonstrate that \method models outperform both their corresponding non-fine-tuned and fine-tuned models on other multilingual instruction datasets for text editing, in addition to achieving strong performance on languages unseen during the task-specific fine-tuning. 
By making our data/models publicly available, we hope to help make advances in multilingual intelligent writing assistants.

\section*{Limitations} 
In this work, we have developed and evaluated instruction-tuned LLMs capable of editing text in multiple languages. However, this work has several limitations that can be improved in future research. 

Despite our attempts to cover a diverse set of seven languages, there are still a number of languages that have not been included in our research, largely because of the lack of high-quality, human-annotated text editing data. Our system can be extended to include more languages in the future to better understand the generalization of these models to new languages, and create more accessible and ubiquitous writing assistants.

Secondly, for training and evaluation, we primarily use datasets that are publicly available in specific languages, and sometimes we generate instructions in English and translate them into multiple languages using Google Translate (for example, for simplification tasks). Despite the fact that our approach allows us to support multiple languages with reasonable development costs, data generated and translated might contain unexpected noise. Moreover, they might not best represent expert-annotated edits in different languages. In order to further improve multilingual LLMs for text editing, future research can use human-generated data for training and evaluation. 

Thirdly, our system leverages numerous LLMs with billions of parameters. Considering the computing resources required for running and developing these models, replicating the results may prove difficult (which we try to address by sharing our models publicly).

Lastly, our evaluations focus only on the performance of the models on benchmark datasets for text editing, which, in turn, focus primarily on measuring superficial characteristics based on n-gram overlaps. These evaluations are limited as they do not test for the more nuanced aspects of text editing, such as fluency, coherence, and meaning preservation. The lack of human evaluations makes assessing these nuanced characteristics difficult. Future work in this direction could look at robust and scalable evaluation metrics for multilingual text editing.

\section*{Ethics Statement}
While our models offer several advantages to make intelligent writing assistance more accessible, we do recognize their potential limitations. Since our work mainly focuses on text editing, we are able to avoid many issues involving generating harmful text. Although there is still a possibility of small meaning changes for stylistic tasks due to the lack of user-specific context \cite{kulkarni2023writing}, we try to reduce the chance of hallucinations by constraining the generation to strictly editing tasks in order to reduce the chance of adding any new information or perpetuating biases.

Moreover, due to the multilingual settings, there is a risk of our models generating responses that are discriminatory, biased, or contain false information. Hence, our models, when fine-tuned on the text editing datasets, may inadvertently learn or propagate these problematic patterns.
To address these concerns and minimize potential harm, we are dedicated to mitigating the risks
associated with the use of our models in future research. We strongly advocate for the responsible use of our models to prevent any unintended negative consequences.

\section*{Acknowledgments}
We sincerely thank Christoph Stuber, David Rojas, Ryan Koo, Samuel Lou, Sofia Mac Gregor, Yuko Okubo, and all the annotators for their immense help, invaluable linguistic expertise, and insightful feedback on the model quality and evaluations. We also thank Max Gubin, Dongyeop Kang, and the anonymous reviewers for their helpful suggestions.

\bibliography{anthology,custom}

\appendix
\clearpage
\section{Training and Evaluation Datasets}
\label{app:training_testing_datasets}

\begin{table*}[t]
  \centering
  \tiny
  \begin{tabular}{p{0.1\textwidth}@{\hskip 1mm}cp{0.3\textwidth}@{\hskip 1mm}p{0.1\textwidth}@{\hskip 2mm}p{0.11\textwidth}}
    \toprule
     \textbf{Task} & \textbf{Language} & \textbf{Dataset} & \textbf{Split} & \textbf{Size} \\
    \midrule
    & & BEA \cite{bryant-etal-2019-bea} & Train  & 1.1M \\
    & en & JFLEG \cite{napoles-etal-2017-jfleg} & Dev, Test & 754, 747 \\
    \cmidrule{2-5}
    & & QALB-2014 \cite{mohit-etal-2014-first} & \multirow{3}{*}{Train, Dev, Test} & 19k, 1k, 968\\
    & ar & QALB-2015 \cite{rozovskaya-etal-2015-second} &  & 310, 154, 920\\
    \makecell[tl]{Grammatical} & & ZAEBUC \cite{habash-palfreyman-2022-zaebuc} &  & 150, 33, 31 \\
    \cmidrule{2-5}
    \makecell[tl]{Error} & de & Falko-MERLIN \cite{boyd-2018-using} & Train, Dev, Test & 19k, 2.5k, 2.3k\\
    \cmidrule{2-5}
    \makecell[tl]{Correction} & es & COWS-L2H \cite{davidson-etal-2020-developing-nlp} & Train, Dev, Test & 398, 85, 86 \\
    \cmidrule{2-5}
    \cmidrule{2-5}
    &  & Lang-8 \cite{mizumoto-etal-2011-mining} & Train & 1.85M \\
    & ja & TEC-JL \cite{koyama-etal-2020-construction} & Test & 1.9k \\
    \cmidrule{2-5}
    & ko & Kor-Union \cite{yoon-etal-2023-towards} & Train, Dev, Test & 108.9k, 23.3k, 23.3k \\
    \cmidrule{2-5}
    & zh & NLPCC-2018 \cite{Zhao2018OverviewOT} & Train, Dev, Test & 540k, 53.5k, 2k \\
    \midrule
    & en & PAWS \cite{zhang-etal-2019-paws} & Train, Dev, Test & 49k, 8k, 8k \\
    \cmidrule{2-5}
    & & SemEval 2017 - Task 1 \cite{cer-etal-2017-semeval} & \multirow{2}{*}{Train, Test} & 1.2k, 67 \\
    & ar & NSURL 2019 - Task 8 \cite{seelawi-etal-2019-nsurl} &  & 24k, 3.7k \\
    & & APB \cite{10.1145/3368691.3368708} & Test & 286 \\
    \cmidrule{2-5}
    Paraphrasing & de & \multirow{10}{*}{PAWS-X \cite{yang-etal-2019-paws}} & \multirow{10}{*}{Train, Dev, Test} & \multirow{10}{*}{49k,	2k, 2k} \\ 
    \cmidrule{2-2}
     & es \\ 
    \cmidrule{2-2}
    & fr \\ 
    \cmidrule{2-2}
    & ja \\ 
    \cmidrule{2-2}
    & ko \\ 
    \cmidrule{2-2}
    & zh \\ 
    \midrule
    & & WikiLarge \cite{zhang-lapata-2017-sentence} & \multirow{2}{*}{Train, Dev, Test} & 296k, 2k, 359 \\
    & en & WikiAuto \cite{jiang-etal-2020-neural} &  & 576k, 5k, 5k \\
    & & NEWSELA \cite{xu-etal-2015-problems} & Train & 94k \\
    & & ASSET \cite{alva-manchego-etal-2020-asset} & Dev, Test & 2000, 359 \\
    \cmidrule{2-5}
    & ar & NEWSELA-Auto-AR & Train & 94k \\
    & & ASSET-Auto-AR & Dev, Test & 100, 359 \\
    \cmidrule{2-5}
    & de & GEOLino \cite{mallinson-etal-2020-zero} & \multirow{2}{*}{Train, Dev, Test} &  958, 122, 118 \\
    & & TextComplexityDE \cite{seiffe-etal-2022-subjective} &  & 200, 25, 25\\
    \cmidrule{2-5}
    Simplification & es & NEWSELA-Auto-ES & Train & 94k \\
    & & ASSET-Auto-ES & Dev, Test & 100, 359 \\
    \cmidrule{2-5}
    \cmidrule{2-5}
    & ja & EasyJapanese \cite{maruyama-yamamoto-2018-simplified} & Train, Dev, Test & 48k, 1k, 1k\\
    & & EasyJapanese Extended \cite{katsuta-yamamoto-2018-crowdsourced} & Train, Test & 34k, 731 \\
    \cmidrule{2-5}
    & ko & NEWSELA-Auto-KO & Train & 94k \\
    & & ASSET-Auto-KO & Dev, Test & 100, 359 \\
    \cmidrule{2-5}
    & zh & NEWSELA-Auto-ZH & Train & 94k \\
    & & CSS \cite{yang-etal-2023-new} & Dev, Test & 383, 383 \\
    \bottomrule
  \end{tabular}
  \caption{\textbf{Datasets used to train and evaluate \methodnosp.} With the exceptions of Spanish GEC and German Simplification, every other dataset contains $>$10k examples for all our experiments.}
  \label{tab:data-sources}
\end{table*}

\subsection{Grammatical Error Correction}

\paragraph{Arabic}
We report on three publicly available Arabic GEC datasets. The first two come from the QALB-2014 \cite{mohit-etal-2014-first} and QALB-2015 \cite{rozovskaya-etal-2015-second} shared tasks. The third is the newly created ZAEBUC dataset \cite{habash-palfreyman-2022-zaebuc,alhafni-etal-2023-advancements}. 
QALB-2014 consists of native/L1 user comments from the Aljazeera news website, whereas QALB-2015 consists of essays written by Arabic L2 learners with various levels of proficiency. It is worth noting that the QALB-2015 dataset has two test sets consisting of L1 and L2 data. In this work, we report results on the L1 test set. The ZAEBUC dataset comprises essays written by native Arabic speakers, which were manually corrected. We use the MaxMatch (M\textsuperscript{2}) Scorer  \cite{dahlmeier-ng-2012-better} for the evaluation.

\paragraph{English}

When it comes to English, we use the  Write \& Improve + LOCNESS (W\&I) corpus released in the Building Educational Applications (BEA) shared task on GEC \cite{bryant-etal-2019-bea}. We also use the NAIST Lang-8 corpus \cite{tajiri-etal-2012-tense}, which is one of the largest and most widely used datasets for English GEC. To test our systems, we use the JFLEG \cite{napoles-etal-2017-jfleg} dataset. We use GLEU \cite{napoles-etal-2015-ground,napoles2016gleu} for the evaluation.

\paragraph{German}
For German, we use the Falko-MERLIN corpus \cite{boyd-2018-using}, which consists of sentences written by L2 learners that were manually corrected. We use the MaxMatch (M\textsuperscript{2}) Scorer  \cite{dahlmeier-ng-2012-better} for the evaluation.

\paragraph{Spanish}
For Spanish, we use the publicly available COWS-L2H~\cite{davidson-etal-2020-developing-nlp} dataset. COWS-L2H consists of essays written by Spanish L2 learners at the university level in the United States.  We use ERRANT  \cite{bryant-etal-2017-automatic} for the evaluation.

\paragraph{Chinese}
For Chinese, we use the data that is part of the NLPCC18 shared task \cite{Zhao2018OverviewOT}. The training data used in the shared task was collected from the NAIST Lang-8 corpus \cite{tajiri-etal-2012-tense}, whereas the test data consists of manually corrected sentences written by Chinese L2 learners. We use GLEU \cite{napoles-etal-2015-ground,napoles2016gleu} for the evaluation.

\paragraph{Japanese}
For Japanese, we use the NAIST Lang-8 corpus \cite{mizumoto-etal-2011-mining} to train our systems. For evaluation, we use the Japanese L2 TEC-JL dataset \cite{koyama-etal-2020-construction}. We use GLEU \cite{napoles-etal-2015-ground,napoles2016gleu} for the evaluation.

\paragraph{Korean}
We use the recently created Kor-Union dataset \cite{yoon-etal-2023-towards}. Kor-Union was created by collecting and combining GEC data from various sources. This includes essays written by Korean native/L1 speakers and L2 learners. We use the MaxMatch (M\textsuperscript{2}) Scorer \cite{dahlmeier-ng-2012-better} for the evaluation.  

\subsection{Paraphrasing}

\paragraph{Arabic} For training, we use the Arabic SemEval Paraphrasing (ASEP) corpus, which sourced three existing Arabic semantic similarity datasets released during SemEval 2017 Task 1 \cite{cer-etal-2017-semeval}, consisting of roughly 1100 sentence pairs. 
For our purposes, similar to them, we only keep the sentence pairs with a semantic similarity score $\geq$ 3.25, which leads to 603 pairs. We also inverted the pairs for training, leading to a total of 1.2k training pairs. For evaluation, we use the evaluation dataset that was used for SemEval 2017 Track 1, but with the same similarity threshold as the training data, consisting of 67 sentence pairs. This evaluation set consists of sentences from the SNLI Corpus \cite{bowman-etal-2015-large} that were human-translated into Arabic, provided by CMU-Qatar by native Arabic speakers with strong English skills.

We also source from the Arabic Question Similarity (Shared Task 8) organized at the Workshop on
NLP Solutions for Under-Resourced Languages (NSURL 2019) \cite{seelawi-etal-2019-nsurl}. The dataset was developed by mawdoo, and consists of 12k pairs for training and 3715 for testing. For both training and evaluation, we filter the semantically similar pairs (similarity score of 1), which leaves us with 10.7k training and 1.7k test pairs. 

We also use the Arabic Paraphrasing Benchmark (APB) dataset \cite{10.1145/3368691.3368708}, which consists of 1010 Arabic sentence pairs that are collected from different Arabic books. Paraphrasing was performed manually using six transformation procedures (i.e., addition, deletion, expansion, permutation, reduction, and replacement). Similar to other evaluation sets, we only keep the sentence pairs with a semantic similarity score $\geq$ 3.25, which leads to 286 pairs.

\paragraph{English}
Paraphrase Adversaries from Word Scrambling (PAWS) is a dataset that contains pairs of sentences with a high lexical overlap \cite{zhang-etal-2019-paws}. We use the PAWS dataset for training and evaluation.

\paragraph{German, Spanish, Japanese, Korean, Chinese}
We use the Cross-lingual Paraphrase Adversaries from Word Scrambling \cite{yang-etal-2019-paws} dataset (PAWS-X), which was created by translating a subset of the PAWS validation and test sets to six other languages by professional translators. 

\subsection{Simplification}
We draw on a variety of existing text simplification datasets in various languages. Table \ref{tab:data-sources} shows the different simplification datasets we draw on in our work and also outlines the training, development, and test settings. 

A major issue with text simplification is the absence of publicly available, human-annotated, sentence-level parallel corpora for some of the languages we considered, such as Arabic, Spanish, and Korean. Therefore, we addressed this by translating the Text Simplification datasets for English to these three languages, in which the parallel data is absent. 
One potential limitation of this approach could be the poor quality of the translation models, which could negatively impact the overall data quality. Therefore, we use the latest Google Translate API\footnote{\url{https://cloud.google.com/translate/docs/advanced/translating-text-v3}} to construct the translated data, and further verify the quality of the translated text with human annotators (native speakers) for a subset of the data. We chose the Google API since it performed best amongst the other open-source machine translation models and APIs we tested\footnote{\url{https://libretranslate.com/}}\footnote{\url{https://www.deepl.com/translator}}.

\paragraph{English}
For English, we used Wikilarge \cite{zhang-lapata-2017-sentence}, WikiAuto \cite{jiang-etal-2020-neural}, and Newsela \cite{xu-etal-2015-problems} datasets for training. WikiAuto is a neural CRF-aligned corpus of original and simple Wikipedia documents that are automatically aligned to generate sentence pairs, whereas the Newsela \cite{xu-etal-2015-problems} dataset contains automatically aligned sentence pairs from documents that are generated by professional writers at Newsela for various grade levels. For testing, we use ASSET \cite{alva-manchego-etal-2020-asset}, which contains ten high-quality human written simplifications for each of the 2,390 sentences from the TurkCorpus \cite{xu2016optimizing}.

\paragraph{German}
We use the GEOLino \cite{mallinson-etal-2020-zero} and TextComplexityDE \cite{seiffe-etal-2022-subjective} datasets for both training and testing. GEOLinoTest contains text about nature, physics, and people from GeoLino, a children’s magazine that was manually simplified by a German linguist to a five to seven-year-old reading level. TextComplexityDE contains 250 complex sentences from German Wikipedia that native speakers manually simplified.

\paragraph{Japanese}
We use the EasyJapanese \cite{maruyama-yamamoto-2018-simplified} and EasyJapaneseExtended \cite{maruyama-yamamoto-2018-simplified} datasets for training and testing. EasyJapanese contains 50k sentence pairs that were manually created by five students by simplifying text from the Tanaka corpus \cite{tanaka2001compilation}. The EasyJapaneseExtended dataset contains an additional 34.4k sentences from the Tanaka corpus with simplifications crowdsourced.

\paragraph{Arabic, Spanish, Korean}
For Arabic, Spanish and Korean, as there were no publicly available sentence-level parallel datasets available, we translated the English simplification datasets. Specifically, we translated the English Newsela dataset for training and ASSET for testing using the Google Translate API, giving us 94k and 359 examples for training and testing, respectively.

\paragraph{Chinese}
We found no publicly available dataset for training Chinese Simplification. Therefore, we again translated the English Newsela training dataset into Chinese. However, for the testing set, we use the CSS \cite{yang-etal-2023-new} dataset. CSS consists of two human-written simplifications for each of the 383 original sentences from the PFR corpus. \footnote{\url{https://www.heywhale.com/mw/dataset}}

\section{Data Preparation}
For Seq2Seq models, we prepend the task-specific instructions to the input to the encoder for each language, performing a full parameter update on the entire sequence. We construct the CLM datasets by wrapping each example in model-specific instructions\footnote{PolyLM and Bactrian-X follow different prompt templates. Details in Appendix \ref{app:verbalizers}.} computing the loss \textit{only} on the target text; hence, in the Native and Random settings, the model does not optimize for the specific ``translated'' instructions.

We randomly sample 10k examples from the original datasets for each language-task combination and keep the original validation and test tests (see \autoref{tab:data-sources}). We chose this quantity as a balance between computational cost and qualitative performance based on the insights from \cite{raheja2023coedit}. Moreover, in our experiments, we did not find a significant impact of increasing the data quantity per task-language combination beyond 10k.
In the Spanish GEC task, we only have 398 data points, so this portion of our data is considerably smaller than the rest of the languages. Furthermore, in the GEC and Simplification training data for all languages, we reserve 20\% of the set as input-output pairs without any edits to avoid over-corrections by the model.

\section{Results}
\label{app:main_results}
We present the full set of results for our best-performing random-language setting. In addition to the marginally higher performance that all trained models demonstrate in this setting, we choose this as our representative setting due to the fact that it is the most capable setting for our models, allowing them to perform cross-lingual editing. 

We present the results for all trained models on all datasets, as well as the No Edits (Copy) baseline and State-of-the-art LLMs. In the interest of space and interpretability, we skip the detailed results for the untrained baseline as we already show them to massively underperform relative to the trained models. We also compare our results to the previously reported SOTA results across tasks and languages:

\paragraph{Grammatical Error Correction} For GEC, we compare against the following SOTA results: Arabic \cite{alhafni-etal-2023-advancements}, English \cite{zhang-etal-2023-bidirectional}, German \cite{fang-etal-2023-improving}, Korean \cite{yoon-etal-2023-towards}, Spanish \cite{flachs-etal-2021-data}, and Japanese \cite{koyama-etal-2020-construction}. 

\paragraph{Simplification} For simplification, we take the results from the fine-tuned mT5 models from \cite{ryan-etal-2023-revisiting}. They compare multiple settings in their work, such as using data from a single training dataset, from a single language, and from multiple languages. We pick the score from any setting that gives the highest score. Thus, the SARI score for German might be picked from one setting while the BLEU score for German from some other setting, and so on. This ensures that we take their strongest models for each use case. We still see that our models perform better than them in most languages and datasets.

\paragraph{Paraphrasing} 
To paraphrase, most of the related works that utilize the PAWS-X dataset \cite{yang-etal-2019-paws} have used it for paraphrase detection, and not for paraphrase generation. Hence, we are unable to report any comparable SOTA results for this task. 



\begin{table*}
\centering
\fontsize{4}{4}\selectfont
\begin{tabular}{lllcccccclll}
\toprule
\textbf{Model} & \textbf{Version} & \textbf{Size} & \textbf{en}  & \textbf{de}  & \textbf{es}  & \textbf{ja}  & \textbf{ko}  & \textbf{zh} & \multicolumn{3}{c}{\textbf{ar}} \\
\cmidrule{4-12}
  & & & JFLEG & Falko-MERLIN & CowsL2H & TEC-JL & Ko-Union & NLPCC-18 & QALB-14 & QALB-15-L1 & ZAEBUC \\
\midrule
Copy & -- & -- & 40.47 & 35.23 & 31.36 & 49.09 & 48.2 & 56.89 & 1.91 & 0.07 & 0.0 \\
\midrule
GPT3.5 & \texttt{gpt-3.5-turbo0613} & -- & 39.54 & 39.41 & 38.89 & 38.56 & 27.5 & 12.5 & 28.08 & 35.01 & 45.17 \\
GPT4 & \texttt{gpt-4-0613} & -- & 35.43 & 40.23 & 38.23 & 40.55 & 29.34 & 11.05 & 41.78 & 44.55 & 47.9 \\
\midrule
Multilingual SOTA & -- & -- &  61.97 & 76.3 & 57.32 &  73.6  & 31.70 & -- & 79.6 & 80.3 & 83.1\\
\midrule
\multirow{3}{*}{mT5} & \texttt{mt5-large} & 1.2B & 36.28 & 12.98 & 40.89 & 12.98 & 35.17 & 50.76 & 64.76 & 64.93 & 68.56 \\
    & \texttt{mt5-xl} & 3.7B & 40.72 & 40.21 & 52.98 & 26.33 & 36.14 & 52.45 & 68.36 & 68.53 & 68.95 \\
    & \texttt{mt5-xxl} & 13B & 41.56 & 40.41 & 51.4 & 39.68 & 37.11 & 55.56 & 67.83 & 67.31 & 67.23 \\ 
\midrule
\multirow{4}{*}{mT0 / Bloomz} & \texttt{mt0-large} & 1.2B & 38.25 & 32.32 & 43.39 & 9.14 & 24.04 & 51.54 & 64.74 & 64.34 & 69.78 \\
            & \texttt{bloomz-3b} & 3.7B & 7.25	& 4.71 & 31.20 & 11.35 & 21.33& 32.12 & 66.21 & 65.68 & \textbf{76.42} \\
            & \texttt{mt0-xl} & 3.7B & 40.3	& 39.6 & 49.19 &  10.22 & 36.53 & 52.25/ & 67.45 & 67.18 & 65.98 \\
            & \texttt{mt0-xxl} & 13B & 40.65 & 40.75 & 52.31 & 14.14 & 37.06 & 52.96 & 68.15 & 67.84 & 66.85 \\
\midrule
mT0 / Bloomz
            & \texttt{bloomz-7b1-mt} & 7.1B & 30.67 & 9.91 & 33.13 & 10.91 & 24.53 & 30.15 & 68&  66.65& 63.15  \\
(mt) & \texttt{mt0-xxl-mt} & 13B & 37.58 & 41.75 & \textbf{53.23} & 46.35 & 57.54 & 53.67 & 70.19 & 69.95 & 69.6 \\
\midrule
PolyLM & \texttt{polylm-multialpaca-13b}	& 	13B & 38.35 & 35.45 & 46.87 & 43.22 & 22.3 & 57.78 & 54.1 & 51.5 & 51.26 \\
\midrule
\multirow{2}{*}{Bactrian-X}  & \texttt{bx-llama-7b} & 7B & 58.67 & 60.07 & 45.32 & 47.1 & 25.56 & 56.09 & 6.99 & 5.16 & 0.73\\
& \texttt{bx-llama-13b} & 13B & \textbf{59.55} & \textbf{64.11} & 49.0 & \textbf{53.41} & 26.66 & \textbf{67.54} & 6.64 & 4.14 & 13.98 \\
\bottomrule
\end{tabular}

\break \vspace{4pt}
\begin{small}
(a) Grammatical Error Correction
\end{small}
\break

\vspace{10pt}

\begin{tabular}{lllcccccclll}
\toprule
\textbf{Model} & \textbf{Version} & \textbf{Size} & \textbf{en}  & \textbf{de}  & \textbf{es}  & \textbf{ja}  & \textbf{ko}  & \textbf{zh} & \multicolumn{3}{c}{\textbf{ar}} \\
\cmidrule{4-12}
  & & & PAWS & PAWS-X & PAWS-X & PAWS-X & PAWS-X & PAWS-X & NSURL & ASEP & APB \\
\midrule
Copy	& --	& 	--	& 0.0 / 100.0	& 0.0 / 100.0	& 0.0 / 100.0	&	0.0 / 100.0 & 0.0 / 100.0	& 0.0 / 100.0	& 0.0 / 100.0 & 0.0 / 100.0	& 0.0 / 100.0\\
\midrule
GPT3.5	& 	\texttt{gpt-3.5-turbo0613}	&  -- &  40.85 / 98.78  & 41.49 / 97.8	& 45.17 / 97.98	& \textbf{82.97} / 
42.88  & 82.97 / 42.88	& 82.38 / 42.19  &  88.2 / 41.33 & 88.57 / 45.7  & 99.55 / 58.94 \\
GPT-4	& 	\texttt{gpt-4-0613}	& 	--	&  36.84 / 97.2  & 48.5 / 95.96	& 44.09 / 95.85	& \textbf{83.45} / 48.38	&  44.81 / 91.47 & 44.60 / 68.49	& 38.98 / 77.23 & 80.0 / 37.23 & 78.80 / 49.36 \\
\midrule
\multirow{3}{*}{mT5}	& 	\texttt{mt5-large}	& 	1.2B	& 	25.77	/	100.00	& 	39.72	/	97.29	& 	24.83	/	96.54	& 	35.80	/	91.13	& 	28.01	/	88.76	& 	36.79	/	94.87	& 	59.94	/	93.79	& 	38.61	/	86.83	& 	14.81	/	68.25	\\
	& 	\texttt{mt5-xl}	& 	3.7B	& 	27.63	/	100.00	& 	32.99	/	96.31	& 	34.01	/	97.22	& 	43.15	/	91.59	& 	40.44	/	88.58	& 	32.35	/	95.81	& 	74.90	/	93.98	& 	64.51	/	86.83	& 	17.99	/	68.25	\\
	& 	\texttt{mt5-xxl}	& 	13B	& 	46.70	/	100.00	& 	52.52	/	97.78	& 	42.31	/	97.13	& 	50.57	/	92.04	& 	57.68	/	88.85	& 	35.22	/	94.96	& 74.99	/	94.45	& 	71.14	/	86.89	& 	36.98	/	68.25	\\
\midrule																														
\multirow{4}{*}{mT0 / Bloomz}	& 	\texttt{mt0-large}	& 	1.2B	& 	24.49	/	100.00	& 	33.66	/	98.12	& 	20.60	/	98.11	& 	25.09	/	92.32	& 	20.13	/	89.65	& 	23.11	/	95.81	& 10.86	/	95.79	& 	46.10	/	86.89	& 	10.99	/	68.25	\\
	& 	\texttt{bloomz-3b}	& 	3.7B	& 	40.53	/	100.00	& 	33.75	/	98.24	& 	32.77	/	98.21	& 	39.87	/	92.03	& 	40.25	/	89.66	& 	45.92	/	95.84	& 	67.68	/	94.83	& 	65.19	/	86.89	& 	63.95	/	68.25	\\
	& 	\texttt{mt0-xl}	& 	3.7B	& 	26.80	/	100.00	& 	32.56	/	98.28	& 	30.17	/	98.18	& 	46.31	/	92.05	& 	33.68	/	89.62	& 	48.10	/	95.85	&  38.79	/	94.34	& 	65.43	/	86.89	& 	17.26	/	68.25	\\
	& 	\texttt{mt0-xxl}	& 	13B	& 	45.39	/	100.00	& 	42.74	/	98.27	& 	41.66	/	\textbf{98.22}	& 	52.68	/	93.12	& 	47.82	/	89.99	& 	51.98	/	95.78	&  61.79	/	94.89	& 	77.72	/	86.89	& 	35.19	/	68.25	\\
\midrule																														
mT0 / Bloomz & 	\texttt{bloomz-7b1-mt}	& 	7.1B	& 	41.30	/	100.00	& 	43.02	/	97.88	& 	42.23	/	98.12	& 	43.67	/	91.12	& 	48.20	/	89.76	& 	54.58	/	95.82	& 	70.62	/	94.74	& 	70.14	/	86.89	& 	67.37	/	68.25	\\
(mt) & 	\texttt{mt0-xxl-mt}	& 	13B	& 	44.74	/	100.00	& 	42.51	/	98.27	& 	39.55	/	98.21	& 	49.71	/	91.23	& 	46.82	/	89.83	& 	49.01	/	95.82	&  75.77	/	94.44	& 	76.31	/	86.89	& 	50.67	/	68.25	\\
\midrule																														
PolyLM	& 	\texttt{polylm-multialpaca-13b}	& 	13B	& 	47.57	/	100.00	& 	43.15	/	94.74	& 	29.33	/	86.89	& 	35.96	/	68.25	& 	30.22	/	\textbf{98.21}	& 	28.32	/	\textbf{98.27}	& 75.32	/	92.05	&	77.41	/	89.66	&	83.04	/	95.82	\\
\midrule																														
\multirow{2}{*}{Bactrian-X}	& 	\texttt{bx-llama-7b}	& 	7B	& 	51.91	/	\textbf{100.00}	& 	50.07	/	\textbf{98.32}	& 	46.67	/	98.11	& 	55.86	/	93.11	& 	54.88	/	89.66	& 	54.18	/	95.82	& 75.37	/	94.74	& 	94.76	/	86.89	& 	94.60	/	68.25	\\
	& 	\texttt{bx-llama-13b}	& 	13B	& 	\textbf{53.49}	/	100.00	& 	\textbf{51.50}	/	98.19	& 	\textbf{48.91}	/	98.21	& 	58.47	/	\textbf{94.23}	& 	59.88	/	89.66	& 	\textbf{55.22}	/	95.89	& 	76.17	/	94.74	& 	93.04	/	86.89	& 	93.42	/	68.25	\\
\bottomrule
\end{tabular}

\break \vspace{4pt}
\begin{small}
(b) Paraphrasing
\end{small}
\break

\vspace{10pt}

\begin{tabular}{lllcccccccccc}
\toprule
\textbf{Model} & \textbf{Version} & \textbf{Size} & \multicolumn{2}{c}{\textbf{en}}  & \textbf{ar}  & \textbf{es}  & \multicolumn{2}{c}{\textbf{de}}  & \multicolumn{2}{c}{\textbf{ja}}  & \textbf{ko} & \textbf{zh} \\
\cmidrule{4-13}
  &
  &
  & ASSET & WikiAuto & ar-ASSET & es-ASSET & GeoLino & TCDE & EasyJ & EasyJE & ko-ASSET & CSS \\
\midrule

Copy	& 	--	&	--	& 	20.73	/	92.81	& 	20.93	/	45.40	& 	17.91	/	86.75	& 	21.17	/	92.56	& 	27.45	/	69.86	& 	15.42	/	26.77	& 	29.66	/	75.91	& 	22.00	/	48.47	& 	16.45	/	82.32	& 	29.27	/	90.42	\\
\midrule
GPT3.5	& 	\texttt{gpt-3.5-turbo0613}	& 	--	& 	38.69	/	53.53	& 	38.99	/	22.00	& 	36.90	/	20.21	& 	43.17	/	51.85	& 	27.81	/	14.77	& 	38.04	/	10.04	& 	20.35	/	12.01	& 	27.88	/	6.35	& 	38.10	/	18.95	& 	21.82	/	15.23	\\
GPT4	& 	\texttt{gpt-4-0613}	& 	--	& 	39.74	/	46.04	& 	39.64	/	19.55	& 	36.77	/	17.76	& 	40.41	/	35.97	& 	24.28	/	9.05	& 	38.43	/	8.47	& 	15.35	/	5.52	& 	26.32	/	4.96	& 	35.81	/	9.84	& 	18.73	/	9.34 \\
\midrule
Multilingual SOTA & -- & -- &  42.77 / 88.26 & 42.48 / 37.95 & -- / -- & -- / -- & 50.75 / 71.9  & 41.15 / 24.53 & 70.95 / 68.12 & 53.49 / 35.67 & -- / -- & -- / -- \\
\midrule
\multirow{3}{*}{mT5}	& 	\texttt{mt5-large}	& 	1.2B	& 	33.10	/	91.04	& 	37.30	/	43.17	& 	36.60	/	76.04	& 	35.60	/	90.62	& 	40.96	/	59.04	& 	31.49	/	22.86	& 	41.85	/	74.93	& 	31.23	/	49.24	& 	33.08	/	76.20	& 	33.58	/	46.02	\\
	& 	\texttt{mt5-xl}	& 	3.7B	& 	31.01	/	92.13	& 	37.16	/	46.27	& 	38.44	/	78.93	& 	37.82	/	88.01	& 	53.07	/	\textbf{73.12}	& 	31.86	/	\textbf{29.27}	& 	65.78	/	79.19	& 	56.38	/	59.36	& 	36.75	/	73.42	& 	32.63	/	39.03	\\
	& 	\texttt{mt5-xxl}	& 	13B	& 	34.49	/	88.05	& 	40.24	/	40.34	& 	40.07	/	68.41	& 	39.47	/	81.96	& 	52.89	/	71.70	& 	32.68	/	26.44	& 	67.76	/	\textbf{79.81}	& 	60.44	/	61.30	& 	38.33	/	66.93	& 	32.29	/	36.16	\\
\midrule																																													
\multirow{4}{*}{mT0 / Bloomz}	& 	\texttt{mt0-large}	& 	1.2B	& 	30.28	/	\textbf{92.81}	& 	34.08	/	46.62	& 	34.87	/	\textbf{83.66}	& 	35.39	/	\textbf{92.12}	& 	44.82	/	69.98	& 	27.84	/	26.20	& 	40.06	/	75.04	& 	25.84	/	47.18	& 	29.18	/	\textbf{80.14}	& 	34.04	/	53.82	\\
	& 	\texttt{bloomz-3b}	& 	3.7B	& 	20.92	/	60.20	& 	21.21	/	30.50	& 	17.99	/	57.87	& 	21.28	/	66.15	& 	27.88	/	44.05	& 	15.73	/	18.40	& 	30.01	/	19.55	& 	23.01	/	17.98	& 	17.35	/	34.56	& 	29.41	/	69.21	\\
	& 	\texttt{mt0-xl}	& 	3.7B	& 	29.63	/	90.96	& 	34.70	/	46.58	& 	36.99	/	77.40	& 	36.40	/	90.26	& 	47.07	/	68.70	& 	30.69	/	26.90	& 	60.62	/	78.02	& 	50.58	/	54.03	& 	33.71	/	77.04	& 	32.71	/	41.33	\\
	& 	\texttt{mt0-xxl}	& 	13B	& 	32.78	/	91.65	& 	38.06	/	44.90	& 	38.93	/	75.76	& 	39.08	/	85.54	& 	50.93	/	70.65	& 	33.92	/	27.29	& 	68.22	/	79.69	& 	\textbf{61.63}	/	\textbf{62.77}	& 	37.51	/	67.01	& 	31.93	/	33.22	\\
\midrule																																													
mT0 / Bloomz 	& 	\texttt{bloomz-7b1-mt}	& 	7.1B	& 	20.88	/	66.85	& 	21.09	/	34.24	& 	17.98	/	60.62	& 	21.24	/	71.10	& 	27.79	/	47.49	& 	15.60	/	20.43	& 	29.81	/	37.77	& 	22.38	/	27.17	& 	17.20	/	54.66	& 	29.33	/	\textbf{74.98}	\\
(mt)	& 	\texttt{mt0-xxl-mt}	& 	13B	& 	20.92	/	60.20	& 	21.21	/	30.50	& 	17.99	/	57.87	& 	21.28	/	66.15	& 	27.88	/	44.05	& 	15.73	/	18.40	& 	30.01	/	19.55	& 	23.01	/	17.98	& 	17.35	/	34.56	& 	29.41	/	69.21	\\
\midrule																																													
PolyLM	& 	\texttt{polylm-multialpaca-13b}	& 	13B	& 	21.12	/	22.33	& 	21.22	/	11.54	& 	18.00	/	18.23	& 	21.43	/	22.77	& 	27.50	/	13.77	& 	15.49	/	7.94	& 	29.69	/	6.15	& 	22.50	/	4.58	& 	16.86	/	19.84	& 	29.53	/	21.18	\\
\midrule												
\multirow{2}{*}{Bactrian-X}	& 	\texttt{bx-llama-7b}	& 	7B	& 	41.05	/	90.79	& 	43.88	/	46.96	& 	40.76	/	74.33	& 	43.57	/	89.80	& 	56.33	/	64.73	& 	\textbf{41.00}	/	27.42	& 	67.47	/	74.64	& 	55.06	/	48.29	& 	38.68	/	71.97	& 	\textbf{43.30}	/	58.35	\\
	& 	\texttt{bx-llama-13b}	& 	13B	& 	\textbf{41.63}	/	91.63	& 	\textbf{44.19}	/	\textbf{47.36}	& 	\textbf{41.75}	/	73.97	& 	\textbf{44.03}	/	88.59	& 	\textbf{63.77}	/	70.28	& 	40.13	/	25.44	& 	\textbf{68.31}	/	74.88	& 	56.89	/	49.89	& 	\textbf{39.91}	/	70.50	& 	41.12	/	48.11	\\
\bottomrule
\end{tabular}

\break \vspace{4pt}
\begin{small}
(c) Text Simplification
\end{small}
\break

\caption{\textbf{Full set of results on the best-performing setting of 10k random-language-prompted data.} For GEC, we report GLEU or $F_{0.5}$ depending on the metric as described in \autoref{app:training_testing_datasets}.
\label{tab:main_results}
For Paraphrasing, the first quantity is 1~$-$~Self-BLEU and the second one is the accuracy of semantic similarity as calculated by \textit{m}USE (explained in \autoref{sec:evaluation}).
Finally, for Simplification, the first quantity is SARI, and the second one is BLEU.
In terms of models, \texttt{bx-llama-7b} denotes the \texttt{MBZUAI/bactrian-x-llama-7b-merged} checkpoint, and \texttt{bx-llama-7b} denotes the \texttt{MBZUAI/bactrian-x-llama-13b-merged} one.
}
\end{table*}

\section{Task Verbalizers}
\label{app:verbalizers}

We present the full list of our manually curated task-specific verbalizers used
for training and evaluations in \autoref{tab:gec_verbalizers}, \autoref{tab:simplification_verbalizers}, and \autoref{tab:verbalizers}.

\begin{table*}[ht]
    \centering
    \begin{tabular}{p{0.1\linewidth}p{0.9\linewidth}}
    \toprule
    \textbf{Language} & \textbf{Verbalizers} \\
    \toprule
    Arabic & \resizebox{\linewidth}{!}{%
    \tiny%
        \begin{tabularx}{1\textwidth}{>{\raggedleft\arraybackslash}X >{\raggedleft\arraybackslash}X >{\raggedleft\arraybackslash}X}


  \<اصلح القواعد> & \<اصلح القواعد النحوية في هذه الجملة> & \<اصلح القواعد النحوية في الجملة> \\ 
 \<اصلح الأخطاء النحوية> & \<اصلح القواعد النحوية> & \<اصلح جميع الأخطاء النحوية> \\ 
 \<اصلح الأخطاء النحوية في هذه الجملة> & \<اصلح القواعد النحوية للجملة> & \<اصلح التناقضات النحوية في الجملة> \\ 
 \<اجعل الجملة نحوية> & \<اصلح الأخطاء في هذا النص> & \<حدث لإزالة الأخطاء النحوية> \\ 
 \<ازل جميع الأخطاء النحوية من هذا النص> & \<حسن قواعد هذا النص> & \<حسن القواعد> \\ 
 \<حسن القواعد النحوية لهذا النص> & \<حسن القواعد النحوية لهذه الجملة> & \<تحسينات نحوية> \\
 \<ازل الأخطاء النحوية> &  & 
 
    \end{tabularx}} \\
    \midrule
    Chinese & \resizebox{\linewidth}{!}{%
    \tiny%
    \begin{tabularx}{1\textwidth}{*{3}{X}}
 \ZH{修复语法} & \ZH{修复这句话的语法} & \ZH{修复句子中的语法} \\
 \ZH{修复语法错误} & \ZH{修复所有语法错误} & \ZH{修复这句话中的语法错误} \\
 \ZH{修复这句话的语法问题} & \ZH{修复句子的语法} & \ZH{修复句子中的不连贯之处} \\
 \ZH{使句子符合语法} & \ZH{使句子流畅} & \ZH{修复本文中的错误} \\
 \ZH{更新以删除语法错误} & \ZH{删除此文本中的所有语法错误} & \ZH{改进本文的语法} \\
 \ZH{提高语法性} & \ZH{提高文本的语法性} & \ZH{提高这句话的语法性} \\
 \ZH{语法改进} & \ZH{删除语法错误} & 
\end{tabularx}}
    \\
    \midrule
    English & \resizebox{\linewidth}{!}{%
    \tiny%
    \begin{tabularx}{1\textwidth}{*{3}{X}}
          Fix grammar & Fix grammar in this sentence & Fix grammar in the sentence \\
Fix grammar errors & Fix grammatical errors & Fix grammaticality \\
Fix all grammatical errors & Fix grammatical errors in this sentence & Fix grammar errors in this sentence \\
Fix grammatical mistakes in this sentence & Fix grammaticality in this sentence & Fix grammaticality of the sentence \\
Fix disfluencies in the sentence & Make the sentence grammatical & Make the sentence fluent \\
Fix errors in this text & Update to remove grammar errors & Remove all grammatical errors from this text \\
Improve the grammar of this text & Improve the grammaticality & Improve the grammaticality of this text \\
Improve the grammaticality of this sentence & Grammar improvements & Remove grammar mistakes \\
Remove grammatical mistakes & Fix the grammar mistakes & Fix the grammatical mistakes
          \end{tabularx}} \\
    \midrule
    German & \resizebox{\linewidth}{!}{%
    \tiny%
    \begin{tabularx}{1\textwidth}{*{3}{X}}
Grammatik korrigieren & Grammatik in diesem Satz korrigieren & Grammatik im Satz korrigieren \\ 
Grammatikfehler beheben & Alle Grammatikfehler beheben & Grammatikfehler in diesem Satz korrigieren \\ 
Grammatik des Satzes korrigieren & Unstimmigkeiten im Satz beheben & Machen Sie den Satz grammatikalisch korrekt \\ 
Machen Sie den Satz fließend & Fehler in diesem Text beheben & Update zum Entfernen von Grammatikfehlern \\ 
Entfernen Sie alle Grammatikfehler aus diesem Text & Verbessern Sie die Grammatik dieses Textes & Verbessern Sie die Grammatik \\ 
Verbessern Sie die Grammatikalität dieses Textes & Verbessern Sie die Grammatikalität dieses Satzes & Grammatikverbesserungen \\ 
Grammatikfehler entfernen & Beheben Sie die Grammatikfehler
\end{tabularx}} \\
    \midrule
    Japanese & \resizebox{\linewidth}{!}{%
    \tiny%
    \begin{tabularx}{1\textwidth}{*{3}{X}}
 \JA{文法を修正してください} & \JA{この文の文法を修正してください} & \JA{文中の文法を修正してください} \\ 
 \JA{文法エラーを修正してください} & \JA{文法上の誤りを修正してください} & \JA{文法性を修正してください} \\ 
 \JA{文法上の誤りをすべて修正してください} & \JA{この文の文法上の誤りを修正してください} & \JA{この文の文法上の間違いを修正してください} \\ 
 \JA{この文の文法性を修正してください} & \JA{文の文法性を修正してください} & \JA{文の非流ちょう性を修正してください} \\ 
 \JA{文を文法的にしてください} & \JA{文を流暢にしてください} & \JA{このテキストのエラーを修正してください} \\ 
 \JA{文法エラーを削除するために更新してください} & \JA{文法上の間違いを修正してください}  & \JA{このテキストの文法を改善してください} \\ 
 \JA{文法性を改善する} & \JA{このテキストの文法性を改善してください} & \JA{この文の文法性を改善してください} \\ 
 \JA{文法の改善} & \JA{文法の間違いを取り除いてください} & \JA{文法上の間違いを取り除いてください} \\ 
 \JA{文法の間違いを修正してください} & \JA{このテキストから文法上の誤りをすべて削除してください}
\end{tabularx}}
    \\
    \midrule
    Korean & \resizebox{\linewidth}{!}{%
    \tiny%
    \begin{tabularx}{1\textwidth}{*{3}{X}}
 \KO{문법 고쳐} & \KO{이 문장의 문법 고쳐} & \KO{문장의 문법 고쳐} \\ 
 \KO{문법 오류 고쳐} & \KO{모든 문법 오류를 고쳐} & \KO{이 문장의 문법 오류를 고쳐} \\ 
 \KO{문장의 문법을 고쳐} & \KO{문장에서 disflucencies 수정} & \KO{문장을 문법적으로 만드십시오} \\ 
 \KO{문장을 유창하게 만드십시오} & \KO{이 텍스트의 오류를 고쳐} & \KO{문법 오류를 제거하기 위한 업데이트} \\ 
 \KO{이 텍스트의 모든 문법 오류를 제거해} & \KO{이 텍스트의 문법을 향상} & \KO{문법성 향상} \\ 
 \KO{이 텍스트의 문법성을 개선하십시오} & \KO{이 문장의 문법성을 햐상} & \KO{문법 향상} \\ 
 \KO{문법 오류 제거} & \KO{문법적 오류 제거} & \KO{문법 오류 수정}
    \end{tabularx}}
    \\
    \midrule
    Spanish & \resizebox{\linewidth}{!}{%
    \tiny%
    \begin{tabularx}{1\textwidth}{*{3}{X}}
Corregir gramática & Corrige la gramática en esta oración & Arreglar la gramática en la oración \\ 
Corregir errores gramaticales & Corregir la gramaticá & Corregir todos los errores gramaticales \\ 
Corregir errores gramaticales en esta oración & Corrige la gramaticá en esta oración & Corregir la gramaticá de la oración \\ 
Corregir la falta de fluidez en la oración & Haz la oración gramatical & Haz que la oración sea fluida \\ 
Corregir errores en este texto & Actualizar para eliminar errores gramaticales & Eliminar todos los errores gramaticales de este texto \\ 
Mejorar la gramática de este texto & Mejorar la gramaticalidad & Mejorar la gramaticalidad de este texto \\ 
Mejorar la gramaticalidad de esta oración & Mejoras gramaticales & Eliminar errores gramaticales \\ 
Corrige los errores de gramaticá & Corrige los errores gramaticales 
    \end{tabularx}}\\
    \bottomrule
\end{tabular}

\caption{\textbf{Grammatical Error Correction instruction verbalizers}. For every language, we craft 27 GEC-specific instructions, increasing their diversity when the model is trained. For this and subsequent tables, we verify the validity of the instructions with native language speakers (\S~\ref{sec:training_details}).}
    \label{tab:gec_verbalizers}
\end{table*}

\clearpage


\begin{table*}[ht]
    \centering
    \begin{tabular}{p{0.1\linewidth}p{0.9\linewidth}}
    \toprule
    \textbf{Language} & \textbf{Verbalizers} \\
    \toprule
    Arabic & \resizebox{\linewidth}{!}{%
    \tiny%
    \begin{tabularx}{1\textwidth}{>{\raggedleft\arraybackslash}X >{\raggedleft\arraybackslash}X >{\raggedleft\arraybackslash}X}
    

    \<بسط الجملة> & \<بسط هذه الجملة> & \<بسط هذا النص> \\
\<اكتب نسخة أبسط للجملة> & \<أعد كتابة الجملة لتكون أبسط> & \<أعد كتابة هذه الجملة بطريقة أبسط> \\
\<أعد كتابة هذه الجملة من أجل التبسيط> & \<أعد كتابة هذا بصيغة أبسط> & \<اجعل الجملة بسيطة> \\
\<اجعل الجملة أبسط> & \<اجعل هذا النص أقل تعقيدًا> & \<اجعل هذا أبسط> \\
\<بسط> & \<تبسيط> & \<غير إلى صياغة أبسط> \\
\<بسط هذه الفقرة> & \<بسط هذا النص> & \<استخدم صياغة أبسط> \\
\<اجعل هذا أسهل للفهم>

    \end{tabularx}} \\
    \midrule
    Chinese & \resizebox{\linewidth}{!}{%
    \tiny%
    \begin{tabularx}{1\textwidth}{*{3}{X}}
 \ZH{简化句子} & \ZH{简化这句话} & \ZH{简化这段文字} \\ 
 \ZH{为该句子写一个更简单的版本} & \ZH{将句子改写得更简单} & \ZH{用更简单的方式重写这句话} \\ 
 \ZH{为简单起见重写这句话} & \ZH{用更简单的措辞重写这个} & \ZH{让句子变得简单} \\ 
 \ZH{让句子变得更简单} & \ZH{让这段文字不那么复杂} & \ZH{让这件事变得更简单} \\ 
 \ZH{简化} & \ZH{改为更简单的措辞} & \ZH{简化这一段} \\ 
 \ZH{使用更简单的措辞} & \ZH{让这更容易理解} 
 \end{tabularx}}   
 \\
 \midrule
 English & \resizebox{\linewidth}{!}{%
    \tiny%
    \begin{tabularx}{1\textwidth}{*{3}{X}}
          {Simplify the sentence} & {Simplify this sentence} & {Simplify this text} \\
{Write a simpler version for the sentence} & {Rewrite the sentence to be simpler} & {Rewrite this sentence in a simpler manner} \\
{Rewrite this sentence for simplicity} & {Rewrite this with simpler wording} & {Make the sentence simple} \\
{Make the sentence simpler} & {Make this text less complex} & {Make this simpler} \\
{Simplify} & {Simplification} & {Change to simpler wording} \\
{Simplify this paragraph} & {Simplify this text} & {Use simpler wording} \\
{Make this easier to understand}
          \end{tabularx}} \\
    \midrule
    German & \resizebox{\linewidth}{!}{%
    \tiny%
    \begin{tabularx}{1\textwidth}{*{3}{X}}
{Vereinfachen Sie den Satz} & {Vereinfachen Sie diesen Satz} & {Vereinfachen Sie diesen Text} \\
{Schreiben Sie eine einfachere Version des Satzes} & {Formulieren Sie den Satz um, damit er einfacher ist} & {Formulieren Sie diesen Satz einfacher um} \\
{Formulieren Sie diesen Satz der Einfachheit halber um} & {Formulieren Sie dies mit einer einfacheren Formulierung um} & {Machen Sie den Satz einfach} \\
{Machen Sie den Satz einfacher} & {Machen Sie diesen Text weniger komplex} & {Machen Sie es einfacher} \\
{Vereinfachen} & {Vereinfachung} & {Änderung zu einer einfacheren Formulierung} \\
{Vereinfachen Sie diesen Absatz} & {Vereinfachen Sie diesen Text} & {Verwenden Sie einfachere Formulierungen} \\
{Machen Sie es verständlicher}
\end{tabularx}} \\
    \midrule
    Japanese & \resizebox{\linewidth}{!}{%
    \tiny%
    \begin{tabularx}{1\textwidth}{*{3}{X}}
 \JA{文を簡略化してください} & \JA{この文を簡単にしてください} & \JA{このテキストを簡略化してください} \\ 
 \JA{文のより簡単なバージョンを書いてください} & \JA{文をもっと簡単に書き直してください} & \JA{この文をもっと簡単に書き直してください} \\ 
 \JA{この段落を簡略化してください} & \JA{これをもっと簡単な表現で書き直してください} & \JA{文を簡単にしてください} \\ 
 \JA{文をもっと簡単にしてください} & \JA{このテキストをより複雑にしないでください} & \JA{これをもっとシンプルにしてください} \\ 
 \JA{簡略化してください} & \JA{簡略化} & \JA{より簡単な表現に変更してください} \\ 
 \JA{わかりやすくするためにこの文を書き直してください} & \JA{より簡単な表現を使用してください} & \JA{これをもっとわかりやすくしてください} 
        \end{tabularx}}
    \\
    \midrule
    Korean & \resizebox{\linewidth}{!}{%
    \tiny%
    \begin{tabularx}{1\textwidth}{*{3}{X}}
 \KO{문장을 간소화} & \KO{이 문장을 간소화하십시오} & \KO{이 텍스트를 간소화} \\ 
 \KO{문장의 간단한 버전 작성} & \KO{문장을 더 간단하게 다시 쓰세요} & \KO{이 문장을 더 간단한 방식으로 다시 써하십시오} \\ 
 \KO{이 문장을 간소 위해 다시 써} & \KO{간단한 표현으로 다시 작성} & \KO{문장을 간소하게 만드십시오} \\ 
 \KO{이 텍스트를 덜 복잡하게 만들어} & \KO{이것을 더 간단하게 만드십시오} & \KO{간소화} \\ 
 \KO{간소한 문구로 바꿔} & \KO{이 단락을 단순화} & \KO{이 텍스트를 단순화} \\ 
 \KO{간단한 표현을 사용해} & \KO{이해하기 쉽게 만들어}
    \end{tabularx}}
    \\
    \midrule
    Spanish & \resizebox{\linewidth}{!}{%
    \tiny%
    \begin{tabularx}{1\textwidth}{*{3}{X}}
    Simplifica la oración & Simplifica esta oración & Simplificar este texto \\ 
Escribe una versión más simple para la oración & Reescribe la oración para que sea más simple & Reescribe esta oración de una manera más simple \\ 
Reescribe esta oración para simplificarla & Reescribe esto con una redacción más simple & Haz la oración simple \\ 
Hacer la oración más simple & Hacer este texto menos complejo & Haz esto más simple \\ 
Simplificar & Simplificación & Cambiar a una redacción más simple \\ 
Simplificar este párrafo & Usa una redacción más simple & Haz que esto sea más fácil de entender 
    \end{tabularx}}\\
    \bottomrule
\end{tabular}

\caption{\textbf{Simplification instruction verbalizers}. For the simplification task, we generate 19 instructions per language, taking care to not change the meaning of the instruction. For more information see \S~\ref{sec:training_details} and Table~\ref{tab:gec_verbalizers}.}
    \label{tab:simplification_verbalizers}
\end{table*}

\clearpage



\begin{table*}[ht]
    \centering
    \begin{tabular}{p{0.1\linewidth}p{0.9\linewidth}}
    \toprule
    \textbf{Language} & \textbf{Verbalizers} \\
    \toprule
    Arabic & \resizebox{\linewidth}{!}{%
    \tiny%
    \begin{tabularx}{1\textwidth}{>{\raggedleft\arraybackslash}X >{\raggedleft\arraybackslash}X >{\raggedleft\arraybackslash}X}
    

  \<أعد صياغة الجملة> & \<أعد صياغة هذه الجملة> & \<أعد صياغة هذا النص> \\ 
 \<اشرح النص> & \<اكتب إعادة صياغة للجملة> & \<اكتب نسخة معاد صياغتها من الجملة> \\ 
 \<أعد كتابة الجملة بصيغة مختلفة> & \<استخدم صياغة مختلفة> & \<أعد كتابة هذه الجملة> \\ 
 \<أعد كتابة هذا النص> 

    \end{tabularx}} \\
    \midrule
    Chinese & \resizebox{\linewidth}{!}{%
    \tiny%
    \begin{tabularx}{1\textwidth}{*{3}{X}}
 \ZH{解释一下句子} & \ZH{解释一下这句话} & \ZH{解释一下这段文字} \\ 
 \ZH{释义} & \ZH{为这句话写一个解释} & \ZH{写出该句子的释义版本} \\ 
 \ZH{用不同的措辞重写句子} & \ZH{使用不同的措辞} & \ZH{重写这句话} \\ 
 \ZH{改写这句话} & \ZH{重写这段文字} & \ZH{重写此文本} \\ 
 \ZH{改写这段文字} 
 \end{tabularx}} \\
 \midrule
 English & \resizebox{\linewidth}{!}{%
    \tiny%
    \begin{tabularx}{1\textwidth}{*{3}{X}}
          {Paraphrase the sentence} & {Paraphrase this sentence} & {Paraphrase this text} \\
{Paraphrase} & {Write a paraphrase for the sentence} & {Write a paraphrased version of the sentence} \\
{Rewrite the sentence with different wording} & {Use different wording} & {Rewrite this sentence} \\
{Reword this sentence} & {Rephrase this sentence} & {Rewrite this text} \\
{Reword this text} & {Rephrase this text}
          \end{tabularx}} \\
    \midrule
    German & \resizebox{\linewidth}{!}{%
    \tiny%
    \begin{tabularx}{1\textwidth}{*{3}{X}}
Umschreiben Sie den Satz & Umschreiben Sie diesen Satz & Umschreiben Sie diesen Text \\ 
Umschreibung & Schreiben sie eine Umschreibung für den Satz & Schreiben Sie eine paraphrasierte Version des Satzes \\ 
Schreiben Sie den Satz mit einem anderen Wortlaut um & Andere Formulierungen verwenden & Schreiben Sie diesen Satz um \\ 
Formulieren Sie diesen Satz um & Diesen Text umschreiben & Diesen Text umformulieren \\ 
Formulieren Sie diesen Text neu 
\end{tabularx}} \\
    \midrule
    Japanese & \resizebox{\linewidth}{!}{%
    \tiny%
    \begin{tabularx}{1\textwidth}{*{3}{X}}
 \JA{文を言い換えてください} & \JA{この文を言い換えてください} & \JA{このテキストを言い換えてください} \\ 
 \JA{言い換えてください} & \JA{文の言い換えを書いてください} & \JA{文の言い換えバージョンを書いてください} \\ 
 \JA{別の表現で文を書き直してください} & \JA{別の表現を使用してください} & \JA{この文を書き直してください} \\ 
 \JA{この文を言い直してください} & \JA{このテキストを書き直してください}
    \end{tabularx}}
    \\
    \midrule
    Korean & \resizebox{\linewidth}{!}{%
    \tiny%
    \begin{tabularx}{1\textwidth}{*{3}{X}}
 \KO{문장을 의역} & \KO{이 문장을 의역} & \KO{이 텍스트를 의역} \\ 
 \KO{의역} & \KO{문장에 대한 의역 쓰기} & \KO{문장을 의역 버전으로 작성하세요} \\ 
 \KO{문장을 다른 단어로 다시 써} & \KO{다른 문구 사용} & \KO{미 문장을 다시 써} \\ 
 \KO{이 문알을 다른 말로 바꿔} & \KO{이 텍스트 다시 쓰기} & \KO{이 텍스트를 바꾸십시오} \\ 
 \KO{이 텍스트를 다른 말로 바꿔보세요} 
    \end{tabularx}}
    \\
    \midrule
    Spanish & \resizebox{\linewidth}{!}{%
    \tiny%
    \begin{tabularx}{1\textwidth}{*{3}{X}}
Parafrasee la oración & Parafrasee esta oración & Parafrasee este texto \\ 
Paráfrasee & Escribe una paráfrasis de la oración & Escribe una versión parafraseada de la oración \\ 
Reescribe la oración con una redacción diferente & Usar una redacción diferente & Reescribe esta oración \\ 
Reformula esta oración & Reescribe este texto & Reformula este texto
    \end{tabularx}}\\
    \bottomrule
\end{tabular}

\caption{\textbf{Paraphrasing instruction verbalizers}. In this case, we create 14 instructions per language so as not to alter the meaning of the task. For more information see \S~\ref{sec:training_details} and Table~\ref{tab:gec_verbalizers}.}
    \label{tab:verbalizers}
\end{table*}

\clearpage

\section{Human Evaluation Annotation Guidelines}
\label{app:annotation_guidelines}

The human experts were asked to rate the \textbf{fluency}, \textbf{adequacy}, and \textbf{accuracy} separately, following the guidelines in \autoref{fig:human_annotations1}, \autoref{fig:human_annotations2} and \autoref{fig:human_annotations3}, respectively.

\begin{figure}
\mdfdefinestyle{roundbox}{innertopmargin=10pt,innerbottommargin=10pt,roundcorner=5pt}
\begin{mdframed}[style=roundbox]
\small

Given the instruction to edit the text and the text to be edited, rate the output on the following dimensions: \hfill \break

\textbf{Fluency}: is the output valid, free from errors, and like how a native speaker of the language would write? \hfill \break
Please use the following scale for your evaluations along every dimension: \hfill \break
\textbf{Very Good}: The output is of high quality, valid, and correct, like a native speaker. \hfill \break
\textbf{Good}: The output is acceptable with minor errors. \hfill \break
\textbf{Average}: The output is relevant but has significant errors. \hfill \break
\textbf{Bad}: The output is subpar quality. \hfill \break
\textbf{Very Bad}: The output is unusable. \hfill
\end{mdframed}
\caption{\textbf{Annotation guidelines for human evaluations for Fluency.}}
\label{fig:human_annotations1}
\end{figure}

\begin{figure}
\mdfdefinestyle{roundbox}{innertopmargin=10pt,innerbottommargin=10pt,roundcorner=5pt}
\begin{mdframed}[style=roundbox]
\small
Given the instruction to edit the text and the text to be edited, rate the output on the following dimensions: \hfill \break
\textbf{Adequacy}: does the output preserve the meaning of the original sentence? \hfill \break

Please use the following scale for your evaluations along every dimension: \hfill \break
\textbf{Very Good}: The output fully preserves the meaning of the text. \hfill \break
\textbf{Good}: The output is semantically similar to the input with minor errors. \hfill \break
\textbf{Average}: The output is semantically similar to the input but has significant errors. \hfill \break
\textbf{Bad}: The output is barely similar to the input. \hfill \break
\textbf{Very Bad}: The output has opposite meaning to the input. \hfill
\end{mdframed}
\caption{\textbf{Annotation guidelines for human evaluations for Adequacy.}}
\label{fig:human_annotations2}
\end{figure}

\begin{figure}
\mdfdefinestyle{roundbox}{innertopmargin=10pt,innerbottommargin=10pt,roundcorner=5pt}
\begin{mdframed}[style=roundbox]
\small
Given the instruction to edit the text and the text to be edited, rate the output on the following dimensions: \hfill \break

\textbf{Accuracy}: how well do the edits made in the output follow the given instructions? \hfill \break
Please use the following scale for your evaluations along every dimension: \hfill \break
\textbf{Very Good}: The output follows the instructions exactly. \hfill \break
\textbf{Good}: The output generally follows the instructions with minor errors. \hfill \break
\textbf{Average}: The instructions are partially followed but has errors. \hfill \break
\textbf{Bad}: The output did not follow the instructions but did not significantly make the text unusable. \hfill \break
\textbf{Very Bad}: The instructions were completely ignored, and wrong edits were made to make the text unusable. \hfill

\end{mdframed}
\caption{\textbf{Annotation guidelines for human evaluations for Accuracy.}}
\label{fig:human_annotations3}
\end{figure}

\end{document}